\documentclass{article}

\usepackage{PRIMEarxiv}
\usepackage[utf8]{inputenc}     % allow utf-8 input
\usepackage[T1]{fontenc}        % use 8-bit T1 fonts
\usepackage{microtype}          % microtypography
\usepackage{url}                % simple URL typesetting
\usepackage{lipsum}             % dummy text
\usepackage{natbib}             % citations

\usepackage{amsmath}
\usepackage{amsfonts}           % blackboard math symbols
\usepackage{nicefrac}           % compact symbols for 1/2, etc.

\usepackage[table]{xcolor}      
\usepackage{graphicx}
\graphicspath{{media/}}         % set image path
\usepackage{fontawesome5}       % icons

\usepackage{booktabs}           % professional-quality tables
\usepackage{tabularx}
\usepackage{multirow}
\usepackage{mdframed}
\usepackage{placeins}           % provides \FloatBarrier

\usepackage{tikz}
\usetikzlibrary{
    shapes,
    shapes.geometric,
    arrows,
    arrows.meta,
    positioning,
    fit,
    backgrounds,
    shadows
}

\usepackage{fancyhdr}
\definecolor{phase0}{RGB}{255,102,102}
\definecolor{phase1}{RGB}{255,178,102}
\definecolor{phase2}{RGB}{255,255,102}
\definecolor{phase3}{RGB}{178,255,102}
\definecolor{phase4}{RGB}{102,255,178}
\definecolor{inputblue}{RGB}{173,216,230}
\definecolor{taskgreen}{RGB}{144,238,144}
\definecolor{outputpurple}{RGB}{221,160,221}

\definecolor{moderate}{RGB}{255, 193, 7}
\definecolor{high}{RGB}{255, 87, 34}
\definecolor{critical}{RGB}{211, 47, 47}
\definecolor{impact}{RGB}{123, 31, 162}

\usepackage[colorlinks=true,linkcolor=blue,citecolor=blue]{hyperref}
\title{
Multi-view Phase-aware Pedestrian-Vehicle Incident Reasoning Framework with Vision-Language Models

}

\author{
 Hao Zhen, Yunxiang Yang, Jidong J. Yang*\\
  College of Engineering\\
 University of Georgia, Athens, GA, USA\\
  \texttt{\{Hao.Zhen, Yunxiang.Yang, Jidong.Yang\}@uga.edu} \\
  }

\begin{document}
\maketitle

\begin{abstract}
Pedestrian-vehicle incidents remain a critical urban safety challenge, with pedestrians accounting for over 20\% of global traffic fatalities. Although existing video-based systems can detect when incidents occur, they provide little insight into how these events unfold across the distinct cognitive phases of pedestrian behavior. Recent vision-language models (VLMs) have shown strong potential for video understanding, but they remain limited in that they typically process videos in isolation, without explicit temporal structuring or multi-view integration. This paper introduces Multi-view Phase-aware Pedestrian-Vehicle Incident Reasoning (MP-PVIR), a unified framework that systematically processes multi-view video streams into structured diagnostic reports through four stages: (1) event-triggered multi-view video acquisition, (2) pedestrian behavior phase segmentation, (3) phase-specific multi-view reasoning, and (4) hierarchical synthesis and diagnostic reasoning. The framework operationalizes behavioral theory by automatically segmenting incidents into cognitive phases, performing synchronized multi-view analysis within each phase, and synthesizing results into causal chains with targeted prevention strategies. Particularly, two specialized VLMs underpin the MP-PVIR pipeline: TG-VLM for behavioral phase segmentation (mIoU = 0.4881) and PhaVR-VLM for phase-aware multi-view analysis, achieving a captioning score of 33.063 and up to 64.70\% accuracy on question answering. Finally, a designated large language model is used to generate comprehensive reports detailing scene understanding, behavior interpretation, causal reasoning, and prevention recommendations. Evaluation on the Woven Traffic Safety dataset shows that MP-PVIR effectively translates multi-view video data into actionable insights, advancing AI-driven traffic safety analytics for vehicle-infrastructure cooperative systems.

\end{abstract}

\section{Introduction}

Traffic safety remains a critical challenge in modern urban environments, with inevitable interactions between vehicles and Vulnerable Road Users (VRUs), particularly pedestrians, representing scenarios of greatest concern. According to the World Health Organization, over 20\% of traffic fatalities involve pedestrians~\cite{who2023global}, highlighting the urgent need for analytical tools that move beyond mere statistics to provide a deep, causal understanding of these incidents. While the increasing availability of video data from infrastracture-side overhead cameras and onboard vehicle cameras, our ability to automatically interpret these collective data for traffic incidents lags significantly behind. Current systems can detect that an incident occurred or even when an incident occured, but they struggle to explain \textit{why}, leaving a gap between raw pixel data and actionable safety diagnostics.

The challenge of understanding pedestrian safety has historically been approached from two distinct angles: statistical causation modeling and computer vision-based detection. However, a disconnect remains between these domains. As noted in road safety literature, crashes are not random but concentrate in specific spatial-temporal configurations, particularly at intersections involving turning movements and complex controls~\cite{das2021fatal,miranda2011link}. Classical accident causation theories, such as Heinrich’s Domino model~\cite{heinrich1941industrial} and Reason’s Swiss Cheese model~\cite{reason1990human}, conceptualize these incidents not as instantaneous events, but as sequential processes involving latent conditions and unsafe acts. This implies a progression through distinct cognitive and behavioral phases, such as perception, judgment, and action. 

Building on this theoretical foundation, human factors and pedestrian behavior studies have proposed several phase-based frameworks to describe road-user interactions. For instance, Gorrini et al.~\cite{gorrini2017crossing,gorrini2016towards} divided the crossing process into three main phases, approaching, appraising, and crossing, capturing the transition from environment scanning to risk assessment and movement initiation. Other frameworks extend this view to include preparatory and post-decision stages, such as the five-phase taxonomy (pre-recognition, recognition, judgement, action, and avoidance) recently adopted in pedestrian–vehicle interaction datasets~\cite{kong2024wts}. Despite these conceptual advances, operationalizing such sequential behavioral dynamics in quantitative safety analysis remains elusive. Current safety surrogates, such as Time-to-Collision (TTC) and Post-Encroachment Time (PET), reduce complex interactions to single extrema values~\cite{ismail2009automated}, failing to capture the temporal evolution of the encounters or the decision-making process described in behavioral models~\cite{ni2016evaluation,papadimitriou2009critical}.

Simultaneously, the field of vision-based incident detection has evolved from rudimentary binary classification to sophisticated trajectory modeling. Modern deep learning approaches utilize CNNs and object trackers (e.g., YOLO~\cite{khanam2024yolov5}, DeepSORT~\cite{wojke2017simple}) to identify anomalies based on motion irregularities~\cite{fang2024visiontad,zhang2023traffic}. Despite these advances, a fundamental limitation persists: the reliance on single-camera perspectives. As highlighted in recent surveys, single-view systems are inherently vulnerable to occlusion, perspective distortion, and scale variation, critical issues in dense urban intersections~\cite{fang2024visiontad}. While multi-camera systems exist, they have primarily focused on network coverage rather than joint behavioral reasoning~\cite{reulke2007traffic,hsu2020traffic}. Consequently, current vision systems treat incidents as indivisible ``abnormal'' clips rather than deconstructing them into the meaningful behavioral phases necessary for causal understanding.

The field of automated traffic incidents analysis is currently undergoing a fundamental paradigm shift. Research is moving beyond rudimentary object detection~\cite{yu2020bdd100k,yang2024enhancing} and binary event classification~\cite{chakraborty2018freeway,zhang2025language}, which treat traffic incidents as instantaneous points in time. Instead, there is a growing demand for semantic reasoning that views traffic incidents as complex, evolving temporal processes, not just what happened, but also \textit{why} and \textit{how} it unfolded. To truly understand a pedestrian-vehicle incident, one must deconstruct it into a causal chain of cognitive and behavioral phases, ranging from pre-recognition and judgment to action and avoidance. However, existing methodologies often remain fragmented, treating detection and behavioral analysis as isolated sub-tasks rather than components of a unified reasoning process.

Recent advances in Large Video-Language Models (LVLMs) offer a potential bridge between visual analysis and semantic reasoning. Models like Video-LLaMA~\cite{zhang2023video}, Video-ChatGPT~\cite{maazi2024video}, and the Qwen-VL family~\cite{Qwen2.5-VL} demonstrate impressive multimodal reasoning capabilities. However, applying these models directly to pedestrian safety analysis reveals two challenges that existing work has failed to adequately address. First, existing approaches lack the temporal semantic understanding necessary for incident causality analysis. Traffic incidents are not single, isolated events but complex sequences of cognitive and behavioral phases. Understanding why an incident occurred requires deconstructing it into stages, representing distinct decision points that determine the outcome. Current systems, however, process videos as continuous streams without distinguishing these transitions, making it difficult to explain why an incident occurred. Second, most existing large vision-language models (LVLMs) are fundamentally misaligned with multi-view video analysis. Pretrained on single, continuous streams, their attention mechanisms fail to reason across synchronized videos, often resulting in hallucinations or invalid outputs. Existing adaptations like TrafficVLM~\cite{dinh2024trafficvlm} and SeeUnsafe~\cite{zhang2025language} attempt to address this but ultimately resort to processing views independently or aggregating clips by severity, thereby destroying the temporal coherence required for true diagnostic reasoning. 

In this paper, we present the Multi-view Phase-aware Pedestrian-Vehicle Incident Reasoning (MP-PVIR) framework, a novel pipeline that transforms directly from raw multi-view videos into comprehensive diagnostic reports explaining not just what happened, but how and why. To the authors' knowledge, this is one of the first research directly perform the pedestrian-vehicle incident reasoning from multiple video streams to incident reports. Our approach synthesizes behavioral theory with advanced AI to systematically deconstruct incidents through four integrated stages: (1) Event-Triggered Multi-View Video Acquisition; (2) Pedestrian Behavior Phase Segmentation; (3) Phase-Specific Multi-View Reasoning; and (4) Hierarchical Synthesis and Diagnostic Reasoning.

Our technical contribution centers on targeted supervised fine-tuning of the Qwen-2.5-VL architecture to create two specialized models with multi-stream videos from different views. We employ a temporal serialization strategy to process multi-view inputs, enabling the model to learn cross-view correlations via self-attention without complex architectural modifications. The Temporal Grounding VLM (TG-VLM) operationalizes accident causation theory by automatically segmenting videos into five behavioral phases (pre-recognition, recognition, judgment, action, avoidance), achieving temporal localization where baseline models fail entirely. The Phase-aware Video Reasoning VLM (PhaVR-VLM) performs synchronized multi-view analysis within each phase, generating dense captions and answering targeted questions to capture nuanced behavioral cues from complementary perspectives. These outputs feed into a Large Language Model that serves as an expert reasoning agent, synthesizing multi-perspective observations into structured diagnostic reports with causal chains at each behavior phase and prevention recommendations.

Evaluation on the Woven Traffic Safety dataset~\cite{kong2024wts} demonstrates that MP-PVIR successfully addresses the fundamental limitations of existing approaches. Where state-of-the-art VLMs produce no valid outputs for multi-view inputs, our framework achieves meaningful temporal grounding (mIoU = 0.4881) and significant improvements in video understanding (33.063 composite score vs. 30.03 baseline). Most critically, it enables comprehensive cross-view reasoning for complex environmental questions with 54.44\% accuracy and up to 64.70\% accuracy for vehicle view question answering.

This work establishes a new paradigm for AI-empowered traffic safety analysis, moving beyond detection to deliver actionable intelligence. By providing phase-by-phase behavioral analysis from multiple viewpoints, identifying possible causal chains, and generating targeted prevention strategies, MP-PVIR bridges the gap between raw video data and the insights needed to save lives. Our contributions include: (1) the first practical framework unifying phase-based behavioral analysis with multi-view video understanding for incident diagnosis; (2) specialized VLMs that enable synchronized multi-view temporal grounding and reasoning, which capabilities are absent in existing models; (3) a hierarchical synthesis approach leveraging a LLM agent that transforms technical observations into human-interpretable diagnostic reports; and (4) a validated protocol demonstrating the viability of automated, interpretable incident analysis for real-world deployment.

\section{Literature review}
Traffic safety where heterogeneous agents: drivers, pedestrians, cyclists, and micromobility users, converge under turning movements, occlusions, and signal-phase uncertainty. A disproportionate share of severe crashes and pedestrian fatalities occur, where milliseconds of recognition and judgment often determine outcomes. Beyond aggregate statistics, pedestrian safety thus demands incident-level understanding that connects behaviors to context, interactions, and infrastructure. Framing the literature through this lens foregrounds intersections as priority sites for targeted diagnosis and intervention, especially for protecting VRUs.

\subsection{Pedestrian--Vehicle Incident Analysis}
Pedestrian--vehicle incident research shows that crashes are not randomly distributed but concentrate in specific combinations of environment, traffic conditions, and manoeuvres. Trend-mining of national crash databases indicates that a substantial and growing share of fatal pedestrian crashes occur at or near intersections at urban context, with recurring patterns involving turning vehicles, darkness, and complex control types \cite{das2021fatal}. At the facility scale, signalized-intersection studies link collision occurrence to built-environment and exposure factors such as land-use mix, network connectivity, transit supply, and pedestrian activity levels \cite{miranda2011link}. These works establish that pedestrian-vehicle crashes arise from structured interaction scenarios, yet they primarily characterize where and under what conditions crashes occur rather than how individual incidents unfold over time.

To move beyond static spatial associations, it is essential to consider the sequential processes that lead to these events. Classical accident causation frameworks provide a more process-oriented view. Building on theories such as Heinrich’s Domino model \cite{heinrich1941industrial} and Reason’s Swiss Cheese model \cite{reason1990human}, road-safety analyses conceptualize crashes as the outcome of a chain of latent conditions, unsafe acts, and failed defenses \cite{saxena2017analysis}. This perspective implies a progression through cognitive and behavioural stages for both drivers and pedestrians—perceiving hazards, interpreting right-of-way, judging gaps, initiating or withholding movement, and attempting avoidance. However, in practice these frameworks are typically applied qualitatively at the system or aggregate level and rarely yield operational definitions of incident phases that can be mapped onto sensor data, trajectories, or video. Thus, while these models emphasize causality as a process, translating them into measurable behavioural phases remains a challenge. 

A complementary line of work examines pedestrian decision-making and behaviour at finer granularity. Situational and behavioural models of crossing decisions describe how pedestrians interpret constraints imposed by traffic, control devices, and geometry, and how these constraints shape the timing and manner of crossing \cite{cambon2009pedestrian}. Critical surveys of pedestrian behaviour models distinguish route-choice and crossing-behaviour formulations, noting that most existing models capture local decisions (e.g., gap acceptance at a single crosswalk) and often neglect the full sequence from perception through appraisal to action and re-evaluation \cite{papadimitriou2009critical}. These studies highlight cognitive and contextual factors, including attention, risk perception, and perception–reaction time, that are central to incident causation, but remain largely decoupled from automated, incident-level analysis.  

Driven by limitations of sparse and biased collision data, pedestrian-vehicle safety research has increasingly turned to conflict-based analysis using trajectories and surrogate safety measures. Automated video-based systems extract pedestrian and vehicle trajectories and compute indicators such as time-to-collision and post-encroachment time to detect and classify conflicts \cite{ismail2009automated}. Building on this, a theoretical framework based on pedestrian–vehicle interaction patterns shows that conflict indicators have different meanings across patterns (e.g., yielding versus non-yielding interactions), and that safety assessment must account for the temporal evolution of the encounter rather than a single minimum value \cite{ni2016evaluation}. These advances mark an evolution from simple collision detection toward behavioural understanding, but incidents are still often reduced to one conflict event characterized by a small set of extrema, rather than being decomposed into cognitively meaningful phases (pre-recognition, recognition, judgment, action, avoidance). This underscores a key methodological gap: while data-driven methods reveal dynamics, they still lack a structured phase-based lens aligned with human decision processes.

Traffic psychology and interaction studies further analyse the coupled nature of driver–pedestrian behaviour in critical scenarios. Field and interview studies at marked, unsignalised crossings document how pedestrians’ wait/go decisions and drivers’ yield/go responses are shaped by speed, distance, traffic density, visual communication (eye contact, gestures, head movements), and distraction \cite{sucha2017pedestrian}. Simulator experiments under different road environments analyze how drivers adjust speed profiles, braking, and gap acceptance as they approach crossings with varying geometry and traffic conditions \cite{bella2017driver}. Overall, pedestrian crashes can be understood as the outcome of dynamic, bidirectional interactions between road users and their environment. Yet most empirical and modelling studies still describe these interactions at a coarse level (e.g., ``yield'' vs.\ ``no yield'', ``conflict'' vs.\ ``no conflict''), rather than as explicit, phase-based progressions of perception, decision, and action for both pedestrians and drivers. This gap between theoretical recognition of behavioural phases and their operational use in incident analysis motivates the phase-aware, incident-level reasoning framework developed in this paper. 

Overall, the evolution of pedestrian–vehicle safety research from spatial pattern recognition and conceptual causation models to behaviourally informed conflict analysis, highlights a persistent disconnect between cognitive theory and operational modelling. Addressing this gap is essential for developing data-driven, phase-aware frameworks that can interpret the temporal reasoning of pedestrian–vehicle incidents.

\subsection{Traditional Computer Vision for Incident Detection}

Vision-based accident detection has evolved rapidly with the rise of deep learning, producing diverse architectures that treat traffic safety as a pattern-recognition problem over image or video streams \cite{fang2024visiontad}. Early systems typically framed accident detection as binary or multi-class classification on surveillance or dashcam videos, distinguishing ``abnormal/crash'' from ``normal'' scenes. Frame- or clip-level convolutional neural network (CNN) classifiers were trained via transfer learning on synthetic or manually constructed crash examples \cite{tamagusko2022deep,zahid2024datadriven}, while spatio-temporal architectures combining 3D CNNs, optical flow, and recurrent modules improved video-level performance on curated datasets \cite{adewopo2024smart}. Other studies extended single-stream models to multi-task formulations, jointly detecting vehicles, classifying severity, and identifying post-collision fire using cascaded detectors and classifiers \cite{basheer2023realtime}. Production-oriented frameworks further integrated object detection (e.g., Mask R-CNN or YOLO-based models) with tracking and kinematic rules (speed drops, overlapping trajectories) to generate real-time alarms from roadside CCTV feeds \cite{ijjina2019computervision,aboah202xvision}. Data scarcity for rare crash events has also motivated sample-efficient training; co-supervised learning with conditional GANs augments sparse positives and stabilizes supervision for incident classifiers \cite{zhen2022co}.

Beyond direct classification, later approaches incorporated motion and trajectory cues to capture abnormal interactions. These methods combine deep object detection with multi-object tracking to derive higher-level spatio-temporal descriptors to capture abnormalities. For instance, Zhang and Sung \cite{zhang2023traffic} used YOLOv5 \cite{khanam2024yolov5} and Deep SORT \cite{wojke2017simple} to extract vehicle trajectories, transform them into influence maps, and detect anomalies through CNNs. Other pipelines synthesized accident trajectories or detected motion irregularities (e.g., abrupt speed or direction changes after impact) for anomaly recognition \cite{tamagusko2022deep,zahid2024datadriven}. Meanwhile, crash-report mining introduced contextual priors by guiding models toward high-risk configurations (e.g., certain manoeuvres or time-of-day patterns) \cite{li2021attention}. These works represent a shift from static visual recognition to motion-aware and context-informed modeling, yet most still summarize each video clip with a single accident/no-accident label rather than capturing the sequence of behavioural phases.

A central limitation of most vision-based systems lies in their reliance on a single camera viewpoint. Single-view surveillance and dashcam videos are vulnerable to occlusion, perspective distortion, and scale variation, especially at intersections where turning vehicles, pedestrians, and roadside infrastructure overlap \cite{fang2024visiontad}. Consequently, existing models often treat each clip as an indivisible event and fail to distinguish the cognitive and behavioural phases of pedestrian behavior when crash happening, such as pre-recognition, recognition, judgment, action, and avoidance. Although surrogate indicators such as time-to-collision are sometimes computed implicitly, they are not embedded in a semantic understanding of the interaction. Multi-camera traffic-monitoring systems have meanwhile focused on network coverage and computational efficiency rather than behaviour-level reasoning \cite{reulke2007traffic, hsu2020traffic}. Edge–cloud architectures and smart-city platforms primarily aim to distribute deep models across camera networks for real-time detection and system scalability \cite{liu2022smart,adewopo2024smart}. Even with synchronized multi-camera data and rich behavioural cues such as pedestrian gaze and head orientation \cite{kong2024wts}, most systems still process each view independently or use multiple cameras only for coverage, without cross-perspective alignment of trajectories or behaviours.

Collectively, traditional computer-vision approaches to traffic safety have progressed from static image/video classification to motion-based anomaly detection and distributed multi-camera architectures. Yet across these developments, crashes are still modeled as single, scene-level events rather than as phase-based progressions of interacting behaviors observed from multiple visual viewpoints, and thus remain incapable of performing true semantic reasoning. Addressing these gaps will require pairing infrastructure-aware sensing (optimal camera/sensor placement) with sample-efficient learning for rare events and models that align evidence across views and over time \cite{yang2023strategic,yang2024information}. These limitations highlight the need to move beyond purely visual classifiers toward models that couple video with language, enabling richer, interpretable, and phase-aware reasoning about pedestrian–vehicle incidents.

\subsection{Large Vision-Language Models for Event Understanding}
The rapid emergence of large foundation models (LFMs) has reshaped artificial intelligence and broadened its impact across engineering applications. In the transportation safety domain, our recent studies have demonstrated how these models can be specialized and aligned for interpretable crash analysis \cite{zhen2025tab, zhen2024leveraging,ZHEN2025100030, zhen2025bridging,zhao2025auto}. These efforts collectively illustrate the transformative potential of foundation models for safety-critical decision understanding and set the stage for extending language reasoning to the visual domain.

Building upon these advances, LVLMs extend language models with visual encoders to jointly interpret images and videos through natural language. They have recently been adapted to video understanding. General video VLMs can answer open-ended questions, generate captions, and perform visual grounding, but their temporal modelling remains coarse. Models such as VideoGLaMM \cite{munasinghe2025videoglamm}, Grounded-VideoLLM \cite{wang2025groundedvideollm}, and ChatVTG \cite{qu2024chatvtg} improve fine-grained spatial or temporal alignment by combining dual-stream encoders, grounding-specific adapters, or multi-stage training schemes, and can localize actions or objects within untrimmed videos \cite{munasinghe2025videoglamm,wang2025groundedvideollm,qu2024chatvtg}. However, these models are trained on single video streams, assume a fixed temporal sampling strategy, and are typically evaluated on generic benchmarks. As a result, they do not capture domain-specific phase structures in safety-critical events or reason jointly across synchronized camera views.

Recent work has begun to explore VLMs for traffic safety and urban scene understanding. CityLLaVA fine-tunes a generic VLM for city scenarios using bounding-box guided visual prompts, task-specific question--answer templates, and efficient block expansion, achieving strong performance on traffic safety description tasks on the WTS dataset \cite{duan2024cityllava,kong2024wts}. TrafficVLM reformulates the Traffic Safety Description and Analysis task as temporal localization with dense captioning, using a video-language backbone to produce long, phase-wise narratives of vehicle- and pedestrian-related events \cite{dinh2024trafficvlm}. SeeUnsafe integrates multimodal large language model agents into a traffic accident analysis pipeline, using severity-based aggregation of clip-level outputs to perform accident-aware classification and visual grounding on WTS data \cite{zhang2025seeunsafe,kong2024wts}. 
A complementary line of work adopts VLMs as decision-support tools: for example, a roadside-focused framework uses GPT-4o with carefully designed prompts and standardized JSON checklists to perceive pedestrian crossing scenarios in a simulator and to support collision-risk prediction and decision-making at crosswalks \cite{teng2025vlm_crossing}. Collectively, these domain-specific adaptations show that VLMs can produce richer, semantically grounded descriptions than traditional computer-vision pipelines and can be guided toward higher-level traffic-safety concepts such as risk, responsibility, and situational awareness. In this direction, structured prompting with collaborative multi-agent knowledge distillation generates standardized scene and risk annotations and distills them into a compact 3B student VLM for low-resolution traffic videos, enabling edge deployment with interpretable, risk-aware captions \cite{yang2025structured}. 
Likewise, a multi-agent visual--language framework uses a mixture-of-experts and task-specific prompting to improve highway-scene understanding across weather classification, pavement wetness assessment, and congestion detection, demonstrating gains in semantic coverage and reliability for safety-relevant tasks \cite{yang2025multi}. 

Despite these advances, several key limitations persist for incident-level analysis with VLMs. First, existing traffic VLMs operate on a single camera view at a time: multi-camera deployments in datasets such as WTS are used for alternative viewpoints or coverage, but not for synchronized, joint reasoning across views \cite{kong2024wts,duan2024cityllava,dinh2024trafficvlm,zhang2025seeunsafe}. Second, temporal structure is usually represented by coarse event segments or severity labels. Even when models perform dense temporal localization or clip-wise captioning, they do not explicitly decompose incidents into cognitively meaningful phases such as pre-recognition, recognition, judgment, action, and avoidance. Third, recent evaluations of VLMs on pedestrian gestures indicate that current models struggle with subtle nonverbal cues, achieving low agreement with expert labels for instructive traffic gestures \cite{bossen2025gestvlm}. 

In parallel, advances in generic video-LVLM research are improving temporal grounding for general events with more vision cues and segmentation but remain agnostic to the cognitive structure of traffic events. VideoGLaMM focuses on pixel-level, spatio-temporal grounding in dialogue \cite{munasinghe2025videoglamm}; Grounded-VideoLLM introduces discrete temporal tokens and a two-stream architecture to sharpen timestamp prediction and temporal reasoning \cite{wang2025groundedvideollm}; and ChatVTG shows that off-the-shelf video dialogue models can be used in a zero-shot manner to retrieve relevant video moments via caption matching \cite{qu2024chatvtg}. These models improve temporal localization but still lack the ability to map video content onto behavioural phases, integrate multi-view evidence, or infer the causal mechanisms underlying crash events.

To summarize, LVLM-based approaches for traffic understanding show clear potential to move beyond detection and tracking toward semantic and cognitive interpretation. However, they continue to process each video stream independently and reason over time at a relatively coarse granularity. No current framework jointly supports synchronized multi-view processing and phase-aware temporal grounding tailored to pedestrian--vehicle incidents reasoning. This gap motivates our proposed multi-view, phase-aware framework, which integrates LVLM-based temporal grounding with cross-view reasoning to generate structured, interpretable, and incident-level diagnoses.

\section{The MP-PVIR Framework: Multi-view Phase-aware Pedestrian-Vehicle Incident Reasoning Framework}

\subsection{Framework Overview and Design Principles}

\begin{figure}[htbp!]
    \centering
    \includegraphics[width=1\linewidth]{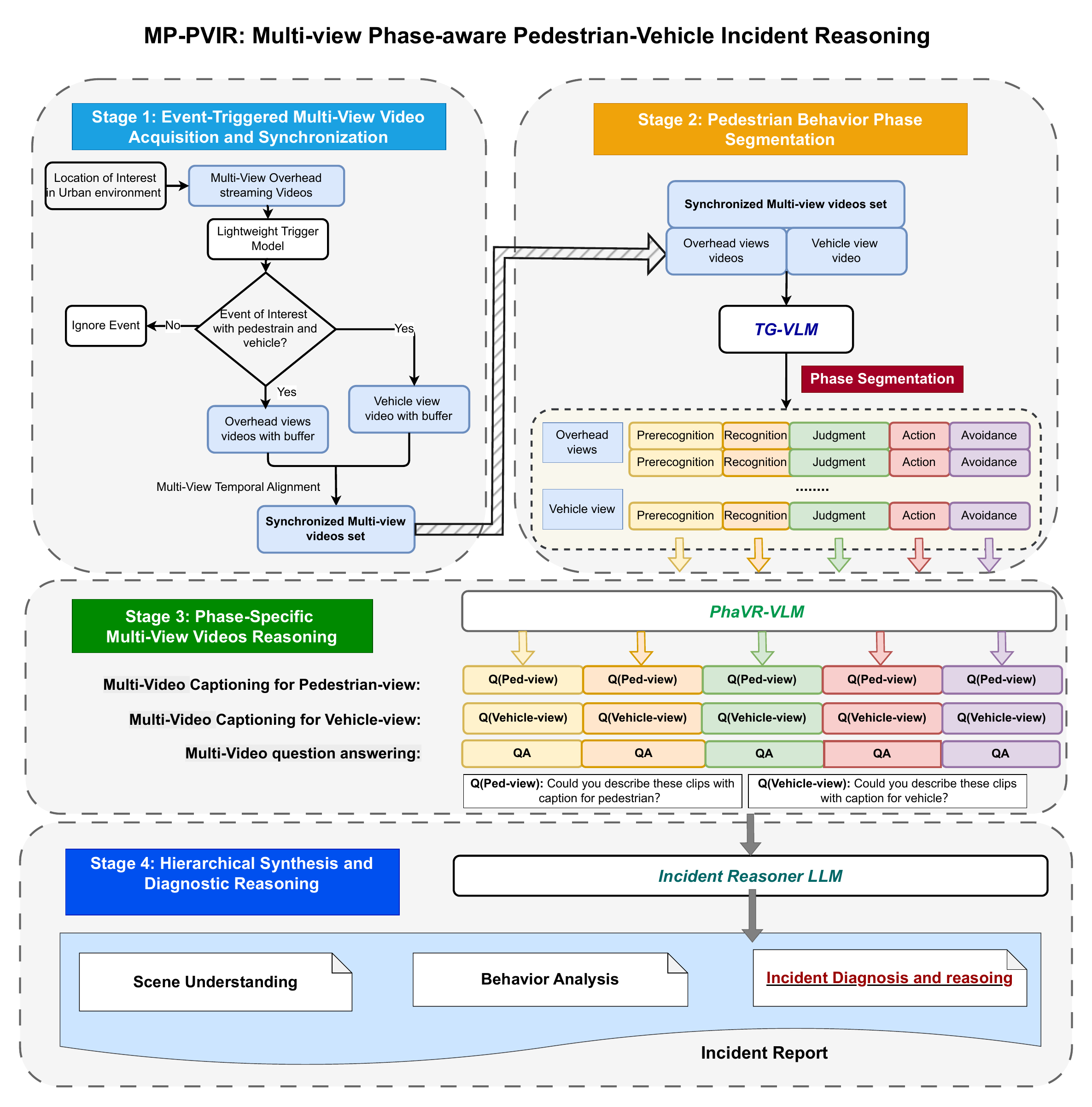}
    \caption{Multi-view Phase-aware Pedestrian-vehicle Incident Reasoning Framework (MP-PVIR).
%Stages~2-4 are implemented and evaluated in this study; Stage~1 is a design-only component relying on standard triggering and synchronization and is not evaluated due to lack of data.
}
    \label{fig:pipeline}
\end{figure}

The Multi-view Phase-aware Pedestrian-vehicle Incident Reasoning (MP-PVIR) framework is a systematic, multi-stage pipeline designed to automatically analyze and generate detailed reports on pedestrian-vehicle interactions from video data. The framework 
operates in four sequential stages: (1) Event-Triggered Multi-View Video Acquisition and Synchronization, (2) Pedestrian Behavior Phase Segmentation, (3) Phase-Specific Multi-View Videos Reasoning, and (4) Hierarchical Synthesis and Diagnostic Reasoning. It is a pipeline designed to process raw multi-view video streams and produce a structured, analytical incident report including the scene understanding, behavior analysis, and event diagnosis and reasoning. The framework moves beyond simple event classification by deconstructing a safety-critical event into its constituent behavioral phases, performing a deep comparative analysis across different viewpoints, and synthesizing these findings into a coherent narrative. %The entire process is designed to be automated, scalable, and capable of delivering actionable intelligence.

\subsection{Stage 1: Event-Triggered Multi-View Video Acquisition and Synchronization}
The process begins with \textit{Stage 1: Event-Triggered Multi-View Video Acquisition and Synchronization}, which is responsible for identifying relevant events and preparing the synchronized multi-view video data. The framework monitors multi-view overhead streaming videos from a location of interest, such as an intersection at an urban environment with high pedestrian volume. To ensure computational efficiency, a ``Lightweight Trigger Model'' continuously processes these streams to detect a predefined ``Event of Interest'', specifically an interaction involving both a pedestrian and a vehicle. It could be defined by the proximity of pedestrian and vehicles along with the detected velocity, flagging potential safety-critical events for further analysis. Upon a positive detection, the system retrieves the relevant segments from buffered videos within a look-back window from overhead and vehicle view cameras. These clips are then passed to a ``Multi-View Synchronization'' module, which ensures temporal alignment between the different video sources. A fundamental prerequisite for any meaningful multi-view fusion is the establishment of a common temporal reference frame. Without precise synchronization, comparing observations from different cameras is meaningless. The output of this stage is a set of video clips from all viewpoints, all precisely aligned to the same event timeline. In our experiments, the WTS dataset already provides synchronized multi-view streams with timestamp alignment~\cite{kong2024wts}; when such annotations are unavailable, common approaches such as motion-energy cross-correlation or feature-based matching can be applied to establish a shared temporal reference. 

\subsection{Stage 2: Pedestrian Behavior Phase Segmentation}

Next, in \textit{Stage 2: Pedestrian Behavior Phase Segmentation}, the synchronized video set is segmented into distinct behavioral phases. Instead of analyzing the video as a monolithic block, MP-PVIR deconstructs it into discrete, semantically meaningful stages based on established models of human behavior in traffic scenarios. The set, containing both overhead and vehicle view videos, is fed into the TG-VLM (Temporal Grounding - Video-Language Model). This model is tasked with performing a "Multi-view videos temporal grounding task," where its goal is to identify and delineate consecutive phases of the pedestrian behavior phase interaction as they appear in the videos. We fine-tuned a Large Temporal Grounding Video-Language Model (TG-VLM), specifically trained for this temporal grounding task, to localize the start and end times for five key behavioral phases: Prerecognition, Recognition, Judgment, Action, and Avoidance. The definitions of these phases follow the WTS dataset~\cite{kong2024wts}, where each incident is annotated by trained experts into semantically distinct stages. Our model learns to detect these expert-labeled boundaries automatically from the video set. This segmentation is performed for all viewpoints, providing a structured table of contents for the incident that guides the subsequent analysis.

\subsubsection{Problem Formulation for Stage 2: Pedestrian Behavior Phase Segmentation}
\label{sec:problem_definition_stage2}
The task of pedestrian behavior phase segmentation can be formulated as a multi-view temporal grounding problem, a task designed to deconstruct a complex event with multiple synchronized videos from different views into a sequence of meaningful behavioral phases. The core idea is to segment a continuous timeline based on high-level natural language descriptions.

A Large Video-Language Model (LVLM) is provided with two main inputs:
\begin{itemize}
    \item Multi-view Video Set ($\mathcal{V}$): A set of $N$ synchronized video streams, $\mathcal{V} = \{V_1, V_2, \dots, V_N\}$, that capture a single event from different perspectives. Each video shares a common duration of $T_{\text{duration}}$. The multi-view nature is crucial as it provides a holistic understanding of the event, helping to resolve occlusions and ambiguities that might be present in any single viewpoint.
    \item Sequential Phase Descriptions ($\mathcal{D}$): An ordered set of $K$ natural language descriptions, $\mathcal{D} = \{D_1, \dots, D_K\}$. Each description $D_k$ defines a specific, semantically distinct behavioral phase $p_k$. In this paper, $K=5$, we focus on the pedestrian behavior phases $p_1$ - $p_5$: pre-recognition, recognition, judgment, action, and avoidance. The sequence of descriptions provides a narrative template for the event's progression, guiding the model's segmentation process.
\end{itemize}

The primary objective of this task is to learn a mapping function $f_{\theta}$, parameterized by the weights $\theta$ of our given LVLM. This function takes $\mathcal{V}$ and $\mathcal{D}$ as input and outputs a set of temporal boundaries $\mathcal{B}$ that delineate the specified five pedestrian behavior phases $p_1$ - $p_5$. The mapping is formally expressed as:\\
\begin{equation}
f_{\theta}: (\mathcal{V}, \mathcal{D}) \rightarrow \mathcal{B}
\end{equation}\\
The output is a set of $K$ temporal boundaries, $\mathcal{B} = \{b_1, b_2, \dots, b_K\}$. Each boundary $b_k$ is a tuple $b_k = (t_k^{\text{start}}, t_k^{\text{end}})$ representing the precise start and end timestamps of phase $p_k$. These timestamps are constrained such that $0 \le t_k^{\text{start}} < t_k^{\text{end}} \le T_{\text{duration}}$. The resulting set $\mathcal{B}$ effectively creates a structured, time-stamped phase segmentation for the analyzed event, making it amenable to further analysis.

% The fundamental challenge lies in the model's ability to perform complex cross-modal reasoning. It must ground the high-level semantics of each textual description $D_k$ with subtle, and potentially distributed, visual cues across all $N$ video streams simultaneously. During training, the model's parameters $\theta$ are optimized by minimizing a loss function $\mathcal{L}$ that quantifies the discrepancy between the predicted boundaries $\hat{\mathcal{B}} = f_{\theta}(\mathcal{V}, \mathcal{D})$ and the ground-truth boundaries $\mathcal{B}_{\text{gt}}$. This loss typically measures the temporal distance or overlap between the predicted and true segments.

% We fine-tune a Qwen-2.5-VL-7B model specifically for temporal grounding of pedestrian behavior phases. The model takes synchronized multi-view videos and phase descriptions as input, outputting temporal boundaries for each phase.

\subsection{Stage 3: Phase-Specific Multi-View Videos Reasoning}

Following pedestrian behavior segmentation, the framework proceeds to the \textit{Phase-Specific Multi-View Videos Reasoning} stage, which performs a in-depth analysis on each identified phase of multiple views. The segmented clips of the same phase are passed to the PhaVR-VLM (Phase-aware Video Reasoning – Video Language Model) for detailed, phase-specific understanding. This model is optimized for detailed video understanding and is tasked with performing two tasks for each view: generating dense, descriptive video captions and answering a set of specific, probing questions (VQA) about the scene and actor behaviors. This process generates a rich, multi-perspective understanding that captures nuanced and often complementary information (e.g., an overhead view may reveal pedestrian distraction while a vehicle view shows a driver's occluded line of sight). The collective outputs of the PhaVR-VLM form a rich, structured set of textual data that describes the entire event chronologically and from multiple viewpoints.

\subsubsection{Problem Formulation for Stage 3: Phase-Specific Multi-View Videos Reasoning}
\label{sec:problem_definition_stage3}
Following the temporal segmentation of the event into distinct phases, this stage performs a detailed analysis and understanding of each phase across all available viewpoints. This process is formulated as two parallel tasks together simultaneously: dense video captioning and visual question answering.

For each identified behavioral phase $p_k$ (where $k \in \{1, \dots, K\}$, $K=5$), the inputs are:
\begin{itemize}
    \item Phase-Segmented Video Clips ($\mathcal{V}_k$): A set of $N$ synchronized video clips, $\mathcal{V}_k = \{V_{1,k}, V_{i,k}, \dots, V_{N,k}\}$. Each clip $V_{i,k}$ corresponds to the video segment from the $i \in \{1, \dots, N=4\}$-th viewpoint, isolated by the temporal boundaries $b_k = (t_k^{\text{start}}, t_k^{\text{end}})$ determined in the previous stage.
    \item Probing Questions ($\mathcal{Q}_k$): For the question-answering task, a predefined set of $M$ natural language questions, $\mathcal{Q}_k = \{Q_{k,1}, \dots, Q_{k,M}\}$, is provided. These questions are tailored to phase $p_k$ to elicit specific information about actor behaviors, environmental conditions, and interactions.
\end{itemize}

The goal is to employ a specialized LVLM, the Incident-VLM ($g_{\phi}$), to interpret the visual content of each clip set $\mathcal{V}_k$. This model executes two tasks within the phase:
\begin{enumerate}
    \item Dense Video Captioning: Generate a descriptive text summary $C_{i,k}$ for each phase clip set $\mathcal{V}_k$.\\
    \begin{equation}
    g_{\phi}^{\text{cap}}: \mathcal{V}_k \rightarrow C_{k}
    \end{equation}
    \vspace{1pt}
    \item Visual Question Answering (VQA): Generate a set of answers $\mathcal{A}_{k}$ for the corresponding set of questions $\mathcal{Q}_k$.\\
    \begin{equation}
    g_{\phi}^{\text{VQA}}: (\mathcal{V}_k, \mathcal{Q}_k) \rightarrow \mathcal{A}_{k}
    \end{equation}
    \vspace{1pt}
\end{enumerate}

The ultimate goal of this stage is to build a comprehensive, multi-faceted understanding of the event on a phase-by-phase basis with video clips from different view. %By systematically generating descriptions and answering specific questions from every viewpoint for specific phase, the framework captures complementary and nuanced details (e.g., an overhead view revealing pedestrian distraction while a vehicle view shows an occluded line of sight). The collective textual output from captioning and also question answering, aggregated across all phases, forms a rich, chronological, and multi-perspective dataset that serves as the foundation for final incident analysis and summarization.

\subsection{Stage 4: Hierarchical Synthesis and Diagnostic Reasoning}

The final stage of the MP-PVIR pipeline, \textbf{Hierarchical Synthesis and Diagnostic Reasoning}, is dedicated to cognition and synthesis. The structured, multi-perspective outputs from the previous stage, including the phase timelines and the corresponding sets of captions and Q\&A pairs from each viewpoint, are aggregated and passed to a Large Language Model (LLM). This LLM functions as a reasoning engine, tasked with integrating, comparing, and synthesizing the context from all sources to formulate a holistic diagnosis. It reasons over the temporal progression of behaviors and the differences in perspective to infer potential causality. The final output is a comprehensive incident diagnosis report, a structured document that transforms the raw video data into actionable intelligence. This report contains four key sections: (1) a high-level scene understanding, (2) a detailed multi-perspective scene analysis, (3) an event diagnosis that infers causality, and (4) a summary on the incident reasoning.

\subsubsection{Problem Formulation for Stage 4: Hierarchical Synthesis and Diagnostic Reasoning}

The objective of the final stage is to move beyond mere description and perform synthesis and diagnostic reasoning to understand the whole incident, infer potential causality, and deliver actionable insights about the incident.
The input to this stage is the complete, structured textual dataset aggregated from all previous stages, which we denote as the comprehensive event information set $\mathcal{I}_{\text{event}}$. This set is a tuple containing:
\begin{itemize}
    \item Phase Timelines ($\mathcal{B}$): The set of temporal boundaries $\{b_1, \dots, b_K\}$ that chronologically structure the event.
    \item Multi-view Captions ($\mathcal{C}$): The collection of caption sets $\{\mathcal{C}_1, \dots, \mathcal{C}_K\}$, providing narrative descriptions for each phase from every viewpoint.
    \item Multi-view Q\&A Pairs ($\mathcal{A}$): The collection of answer sets $\{\mathcal{A}_1, \dots, \mathcal{A}_K\}$, containing specific, targeted information extracted from each phase and viewpoint.
\end{itemize}
Collectively, $\mathcal{I}_{\text{event}} = (\mathcal{B}, \mathcal{C}, \mathcal{A})$ represents a rich, multi-faceted chronicle of the event.
The primary objective is to employ a powerful Large Language Model (LLM), which functions as a reasoning engine $h_{\psi}$, to perform hierarchical synthesis. 
The transformation is defined as:

\begin{equation}
    h_{\psi}: \mathcal{I}_{\text{event}} \rightarrow \mathcal{R}
\end{equation}
\\
This model is tasked with integrating and reasoning over the entirety of the provided information set $\mathcal{I}_{\text{event}}$ to produce a final, diagnostic report $\mathcal{R}$. 

\section{Model Architecture and Training Framework}
\subsection{Data}

In this study, we adopt the WTS dataset~\cite{kong2024wts}, a large-scale, pedestrian-centric traffic video benchmark designed for fine-grained spatio-temporal understanding, for our proposed MP-PVIR framework evaluation. WTS contains two subsets: (i) an internal set of over 1,200 staged pedestrian–vehicle events across 255 distinct traffic scenarios, recorded from multiple synchronized viewpoints (including overhead and in-vehicle cameras), and (ii) an external set of about 4,800 pedestrian-centric videos curated from BDD100K \cite{yu2020bdd100k}. Each video is segmented into five critical behavioral phases: pre-recognition, recognition, judgment, action, and avoidance, capturing the temporal evolution of pedestrian–vehicle interactions. Rich textual captions are provided for each phase from both pedestrian and driver perspectives, supporting detailed behavior analysis. For our experiments, we follow the official WTS training and testing splits for our TG-VLM and PhaVR-VLM training for Stage 2 and Stage 3. Stage~2 (pedestrian behavior segmentation) and Stage~3 (phase-specific multi-view reasoning) are evaluated using the expert annotations provided in WTS, where trained annotators label synchronized multi-view streams with phase-level temporal boundaries. These annotations serve as the ground truth against which our models are trained and evaluated.

\subsection{Model selection and adaptation}

To implement the theoretical architecture of the MP-PVIR framework described in Section 3, we require a robust foundation model capable of processing multi-modal inputs. In this study, we select the Qwen-2.5-VL architecture as the unified backbone to operationalize the core reasoning modules of our pipeline for Stage 2 and Stage 3. Specifically, we adapt this foundation model into two specialized variants to address the distinct requirements of the framework's intermediate stages: the \textit{TG-VLM} is developed to execute the \textit{Pedestrian Behavior Phase Segmentation} (Stage 2), while the \textit{PhaVR-VLM} is engineered to perform the \textit{Phase-Specific Multi-View Reasoning} (Stage 3). This section details the architecture of the base model and the supervised fine-tuning strategies employed to tailor it for these domain-specific traffic safety tasks. 

\subsubsection{Large Video Language Model: Qwen-2.5-VL}

%\paragraph{Qwen2.5-VL Architecture and Video Processing.}
Qwen-2.5-VL \cite{Qwen2.5-VL} is a series of advanced vision-language models, designed to seamlessly integrate visual perception with natural language processing. The model's architecture consists of three core components: a Vision Transformer (ViT) that acts as the visual encoder, a Large Language Model (LLM) based on the Qwen-2.5 series that serves as the decoder, and an MLP-based merger to connect them.

A key innovation in Qwen-2.5-VL is its ability to process video inputs natively and efficiently. Instead of standardizing video inputs, it employs dynamic frame rate (FPS) sampling, allowing it to adapt to videos with variable temporal speeds. Furthermore, it introduces Multimodal Rotary Position Embedding (MROPE) aligned to absolute time. This technique enables the model to understand the precise timing of events within long-duration videos, facilitating sophisticated temporal reasoning and event localization down to the second. This is achieved by aligning temporal IDs with absolute timestamps, allowing the model to perceive the tempo of events directly from the data.

\paragraph{Video Temporal Segmentation Task}
Video temporal segmentation involves partitioning continuous video streams into discrete, semantically meaningful temporal segments based on content changes or activity transitions. This task requires models to identify precise boundaries between different events or behaviors within video sequences, enabling structured analysis of complex temporal progressions. Traditional approaches relied on hand-crafted features and rule-based temporal modeling, but recent advances in Large Video-Language Models have enabled more sophisticated methods that can ground natural language descriptions directly in video content. These models can identify temporal boundaries based on semantic descriptions rather than low-level visual cues alone, as demonstrated in recent work on video temporal grounding~\cite{guo2025vtg}. Our work applies this emerging capability to pedestrian behavior analysis in traffic scenarios, specifically segmenting pedestrian-vehicle interactions into five distinct behavioral phases: pre-recognition, recognition, judgment, action, and avoidance. This particular domain of pedestrian safety presents challenging conditions due to the subtle nature of cognitive transitions, brief duration of critical decision-making moments, and the need for precise temporal localization of behaviors that may exhibit minimal observable visual changes while carrying profound safety implications.

\paragraph{Video Understanding Task}
Video understanding is a crucial area of artificial intelligence that aims to enable machines to interpret and reason about dynamic visual content. These tasks go beyond static image analysis to encompass a wide range of capabilities, including: recognizing complex activities, tracking object interactions over time, understanding temporal and causal relationships between events, and answering detailed questions about the video's content. 

\subsubsection{Supervised Fine Tuning for Video Language Model}
To bridge the gap between the general capabilities of the foundational Qwen2.5-VL model and the requirements of our tasks defined in Section \ref{sec:problem_definition_stage2} and Section \ref{sec:problem_definition_stage3}, we employ a comprehensive supervised fine-tuning strategy. This process adapts the pre-trained model to our specific data distribution and task formats, creating two specialized models, TG-VLM and PhaVR-VLM, for Pedestrian Behavior Phase Segmentation and Phase-Specific Multi-View Analysis, respectively. Central to our approach is the use of parameter-efficient fine-tuning (PEFT) to ensure computational feasibility without sacrificing performance.

\paragraph{Multi-View Input Representation}
To enable the model to process multiple synchronized video streams simultaneously within a unified context window, we employ a Temporal Serialization strategy. Given a set of $N$ synchronized video clips $\mathcal{V} = \{V_1, V_2, \dots, V_N\}$ corresponding to the same event, we concatenate them along the temporal dimension to form a single continuous input sequence $V_{input} = [V_1 \oplus V_2 \oplus \dots \oplus V_N]$. Each video clip is converted into a sequence of visual tokens by the Qwen2.5-VL vision encoder. Temporal Serialization therefore produces unified token sequence: 
$[\text{tokens}(V_1)\ ;\ \text{tokens}(V_2)\ ;\ \cdots\ ;\ \text{tokens}(V_N)
]$

During supervised fine-tuning, the model exploits the long-context attention capabilities of the Qwen2.5-VL architecture to learn cross-view correspondences. Through global attention over the serialized token sequence, the fine-tuned LVLM associates semantically aligned high-resolution visual tokens originating from both overhead and vehicle viewpoints, thereby capturing a coherent representation of the same pedestrian–vehicle event.

\paragraph{Parameter-Efficient Fine-Tuning with LoRA.}
Given the immense scale of modern LVLMs, fine-tuning all model parameters is computationally prohibitive and can lead to catastrophic forgetting. To circumvent this, we leverage PEFT, specifically the Low-Rank Adaptation (LoRA) method. LoRA freezes the original pre-trained model weights and injects smaller, trainable rank-decomposition matrices into the layers of the Transformer architecture. For a pre-trained weight matrix $W_0 \in \mathbb{R}^{d \times k}$, its update $\Delta W$ during fine-tuning is constrained by a low-rank decomposition:

\begin{equation}
W = W_0 + \Delta W = W_0 + BA
\end{equation}
\\
where $B \in \mathbb{R}^{d \times r}$ and $A \in \mathbb{R}^{r \times k}$ are the trainable low-rank matrices, and the rank $r \ll \min(d, k)$. This significantly reduces the number of trainable parameters, retaining the powerful base knowledge of the model while efficiently learning task-specific adaptations.

\paragraph{Training Objective and Data Formatting.}
Both specialized models are trained using a standard causal language modeling objective, where the goal is to predict the next token in a sequence given the previous tokens and the multimodal context. The training objective is to minimize the cross-entropy loss $\mathcal{L}_{\text{LM}}$ over a target text sequence $S$:

\begin{equation}
\mathcal{L}_{\text{LM}} = -\sum_{i=1}^{|S|} \log P(s_i | s_{<i}, \mathcal{V}_{\text{context}}; \theta)
\end{equation}
\\
where $s_i$ is the token at position $i$, $s_{<i}$ are the preceding tokens, $\mathcal{V}_{\text{context}}$ represents the input from the multi-view video clips, and $\theta$ are the trainable LoRA parameters.

\section{Stage-Specific Implementations for MP-PVIR}
\subsection{TG-VLM for Stage 2 Pedestrian Behavior Phase Segmentation} 

To address the temporal grounding task, we develop a specialized model, the TG-VLM, by fine-tuning the Qwen-2.5-VL-7B model. This Large Video-Language Model (LVLM) is chosen for its native ability to process multiple video inputs concurrently and its strong instruction-following and text generation capabilities, which are essential for our target tasks. As illustrated in Figure~\ref{fig:pipeline}, this corresponds to Step~2 of the framework, where raw synchronized videos are segmented into different pedestrian behavior phases that serve as anchors for subsequent reasoning.

\subsubsection{Model Input and Prompting}
The core of our method lies in framing the temporal grounding task as a structured question-answering problem. The TG-VLM is provided with all synchronized video streams from the event ($\mathcal{V}$) and a carefully constructed text prompt. The prompt, shown below, instructs the model to act as a traffic event analyst and provides the explicit definitions for the five behavioral phases ($\mathcal{D}$).

\textit{"You are provided with multiple synchronized videos of a traffic event from different viewpoints. Your task is to identify and locate the temporal boundaries for five distinct phases based on the following definitions: [Phase 0 definition], [Phase 1 definition], ..., [Phase 4 definition]. Provide the start and end timestamps for each phase in seconds."}

\textit{[Phase 0 definition]: Phase 0 (Pre-recognition): The timing before the start of environment awareness (crosswalks, traffic signals, vehicles, etc.).}

\textit{[Phase 1 definition]: Phase 1 (Recognition): The timing from the start of environment awareness (crosswalks, traffic signals, vehicles, etc.) until a judgment is made.}

\textit{[Phase 2 definition]: Phase 2 (Judgment): In principle, the moment from which environmental awareness is completed until the start of an action.}

\textit{[Phase 3 definition]: Phase 3 (Action): Start of movement of any part of the body (excluding eyes and ears) up to the time a result (e.g., collision) occurs.}

\textit{[Phase 4 definition]: Phase 4 (Avoidance): The time after avoidability is clear until the time of avoidance happened or failure to avoid.}

\subsubsection{Evaluation Metrics}
Let $j \in \{0, 1, 2, 3, 4\}$ denote the phase index, $M$ represent the total number of samples in the evaluation set, and $P_{j,k}$, $G_{j,k}$ represent the predicted and ground truth temporal intervals for phase $j$ in sample $k$, respectively. The phase-wise mean IoU for phase $j$ is defined as:
\\
\begin{equation}
\text{mIoU}_j = \frac{1}{M} \sum_{k=1}^{M} \text{IoU}(P_{j,k}, G_{j,k})
\end{equation}
\\
where $\text{IoU}(P_{j,k}, G_{j,k})$ represents the Intersection over Union between the predicted and ground truth temporal intervals for phase $j$ in sample $k$.

% This metric is particularly valuable for understanding model behavior across different temporal phases in complex multi-phase temporal grounding tasks.

\subsection{PhaVR-VLM for Stage 3 Phase-Specific Multi-View Videos Reasoning}
This corresponds to Step~3 in Figure~\ref{fig:pipeline}, where each phase is analyzed in detail to produce captions and answers from multiple viewpoints, forming the intermediate evidence that will later be synthesized into a structured incident report in Step~4.

\subsubsection{Model Input and Prompt}
Our training dataset comprises 33,603 multimodal examples derived from distinct data sources within the WTS dataset. The preparation process involved four main components:

For video captioning tasks, we extracted video caption annotations for both pedestrian and vehicle perspectives from events containing both overhead and vehicle view recordings. Each example combines multiple overhead camera feeds with corresponding vehicle-view footage, targeting specific phase segments with human-annotated captions. Additional caption data from annotated external BDD sources is also used to augment training diversity. The textual prompt follows either `Could you describe this video with caption for pedestrian?'' or ``Could you describe this video with caption for vehicle?''.

For question-answering tasks, three QA categories were processed:
\begin{itemize}
    \item \textit{Environment Analysis}: Full-video questions about traffic environment and scene understanding
    \item \textit{Vehicle View QA}: Temporal-segmented questions from vehicle perspective 
    \item \textit{Overhead View QA}: Multi-camera overhead analysis with temporal constraints
\end{itemize}
All QA prompts follow the format: ``[Question] [Multiple Choice Options] Please output your final answer after `answer\_choice': .''

The data preparation pipeline automatically matched video files with annotations, extracted temporal segments, and formatted conversations into a standardized message structure. Each training example follows a multi-turn conversation format with video inputs and corresponding text responses.

\subsubsection{Evaluation Metrics}
% (1) Video captioning task

To measure the accuracy of a candidate caption compared to the original caption, the evaluation system uses four metrics: BLEU, Rouge-L, Meteor, and CIDEr. The final score is computed as the average of these four metrics:

\begin{equation}
    \text{Score} = \frac{\text{BLEU} + \text{Rouge-L} + \text{Meteor} + 0.1*\text{CIDEr}}{4}
\end{equation}
\\

1) BLEU\\

BLEU provides an automatic and quantitative estimate of the quality of the candidate sentence compared to the target sentence. The formula to calculate the BLEU score is:

\begin{equation}
    \text{BLEU} = \text{BP} \times \exp\left(\sum_{n=1}^{N_B} w_n \cdot \log\left(\frac{\text{clipped\_count}_n}{\text{count}_n}\right)\right)
\end{equation}
\\
where BP represents the brevity penalty, calculated as:

\begin{equation}
    \text{BP} = 
    \begin{cases} 
      1 & \text{if } c_l > r_l \\
      e^{(1 - r_l/c_l)} & \text{if } c_l \leq r_l
   \end{cases}
\end{equation}
\\
$N_B$ denotes the maximum n-gram order (typically 4), $w_n$ is the weight for n-gram order $n$ (typically uniform, $1/N_B$), $\text{clipped\_count}_n$ is the total count of n-grams in the candidate sentence that appear in the target sentence (clipped by the maximum count in the target), $\text{count}_n$ is the total count of n-grams in the candidate sentence, and $c_l$ and $r_l$ are the lengths of the candidate and reference sentences, respectively.\\

2) Rouge-L\\

The Rouge-L metric evaluates the longest common subsequence (LCS) between the generated and reference descriptions. It is formulated as:

\begin{equation}
    \text{Rouge-L} = \frac{(1 + \beta^2) R_{lcs} P_{lcs}}{R_{lcs} + \beta^2 P_{lcs}}
\end{equation}
\\
where,\\
\begin{equation}
    P_{lcs} = \frac{\text{LCS}(X, Y)}{c_l}
\end{equation}
\\
\begin{equation}
    R_{lcs} = \frac{\text{LCS}(X, Y)}{r_l}
\end{equation}\\
$\text{LCS}(X, Y)$ is the length of the longest common subsequence between the reference string X (length $r_l$) and the candidate string Y (length $c_l$). $\beta$ is typically set to 1.\\

3) Meteor\\

The Meteor metric extends beyond simple unigram matching by considering synonyms, stemming, and word order. It is calculated as follows:\\
\begin{equation}
    \text{Meteor} = F_{mean} \times (1 - \text{Penalty})
\end{equation}\\
where,\\
\begin{equation}
    \text{Penalty} = 0.5 \left( \frac{\text{chunks}}{u_m} \right)^3
\end{equation}\\
\begin{equation}
    F_{mean} = \frac{10PR}{R + 9P}
\end{equation}\\
\begin{equation}
    P = \frac{u_m}{c_m} \quad \text{and} \quad R = \frac{u_m}{r_m}
\end{equation}\\
In these formulas, `chunks` is the number of matching unigram chunks, $u_m$ is the number of mapped unigrams between the reference and candidate, and $c_m$ and $r_m$ are the unigram counts in the candidate and reference sentences, respectively.\\

4) CIDEr\\

The CIDEr (Consensus-based Image Description Evaluation) metric quantifies the coherence between a generated caption and a set of human-written reference captions by performing a TF-IDF weighting for each n-gram.
\begin{equation}
    \text{CIDEr}(c_i, S_i) = \sum_{n=1}^{N_C} w_n \text{CIDEr}_n(c_i, S_i)
\end{equation}
where,\\
\begin{equation}
    \text{CIDEr}_n(c_i, S_i) = \frac{1}{m} \sum_{j} \frac{g^n(c_i) \cdot g^n(s_{ij})}{||g^n(c_i)|| \cdot ||g^n(s_{ij})||}
\end{equation}\\
The TF-IDF weighting $g_k(s_{ij})$ for an n-gram $\omega_k$ is:\\
\begin{equation}
    g_k(s_{ij}) = \frac{h_k(s_{ij})}{\sum_{\omega_l \in \Omega} h_l(s_{ij})} \log\left(\frac{|I|}{\sum_{I_p \in I} \min(1, \sum_q h_k(s_{pq}))}\right)
\end{equation}\\
$g^n(c_i)$ is the vector of TF-IDF weights for all n-grams in the candidate sentence $c_i$, and $S_i = \{s_{i1}, ..., s_{im}\}$ is the set of reference captions.\\

% (2) Question answering task

For the multiple-choice question answering evaluation, we compute accuracy as the percentage of correctly predicted choices. Given a dataset of $N$ examples, where each example has a ground truth answer choice $y_i \in \{a, b, c, d\}$ and a predicted choice $\hat{y}_i$ extracted from the model's generated response, accuracy is defined as:

\begin{equation}
\text{Accuracy} = \frac{1}{N} \sum_{i=1}^{N} \mathbf{1}[y_i = \hat{y}_i] \times 100\%
\end{equation}\\
where, $\mathbf{1}[\cdot]$ is the indicator function that equals 1 when the condition is true and 0 otherwise. \\

We employ pattern-matching techniques to extract the choice letter from the model's free-form text responses, using regular expressions to identify patterns such as ``answer: a'', ``choice: b'', or standalone letters. Additionally, we report the \textit{valid choice rate}, which measures the percentage of responses that contain extractable valid choices (a, b, c, or d), providing insight into the model's adherence to the multiple-choice format.

\subsection{Implementation for Stage 4 Hierarchical Synthesis and Diagnostic Reasoning}
\subsubsection{LLM Selection and Configuration}
For the final synthesis stage, we employ Claude Opus 4 \cite{claude_opus_4_2025} as our reasoning engine due to its capabilities in complex multi-document synthesis and structured output generation. The model receives the complete structured dataset from previous stages, including event temporal boundaries, phase-specific captions from multiple viewpoints, and comprehensive Q\&A pairs.

\subsubsection{Prompt Engineering}

\begin{figure}[htbp!]
    \centering
    \includegraphics[width=1\linewidth]{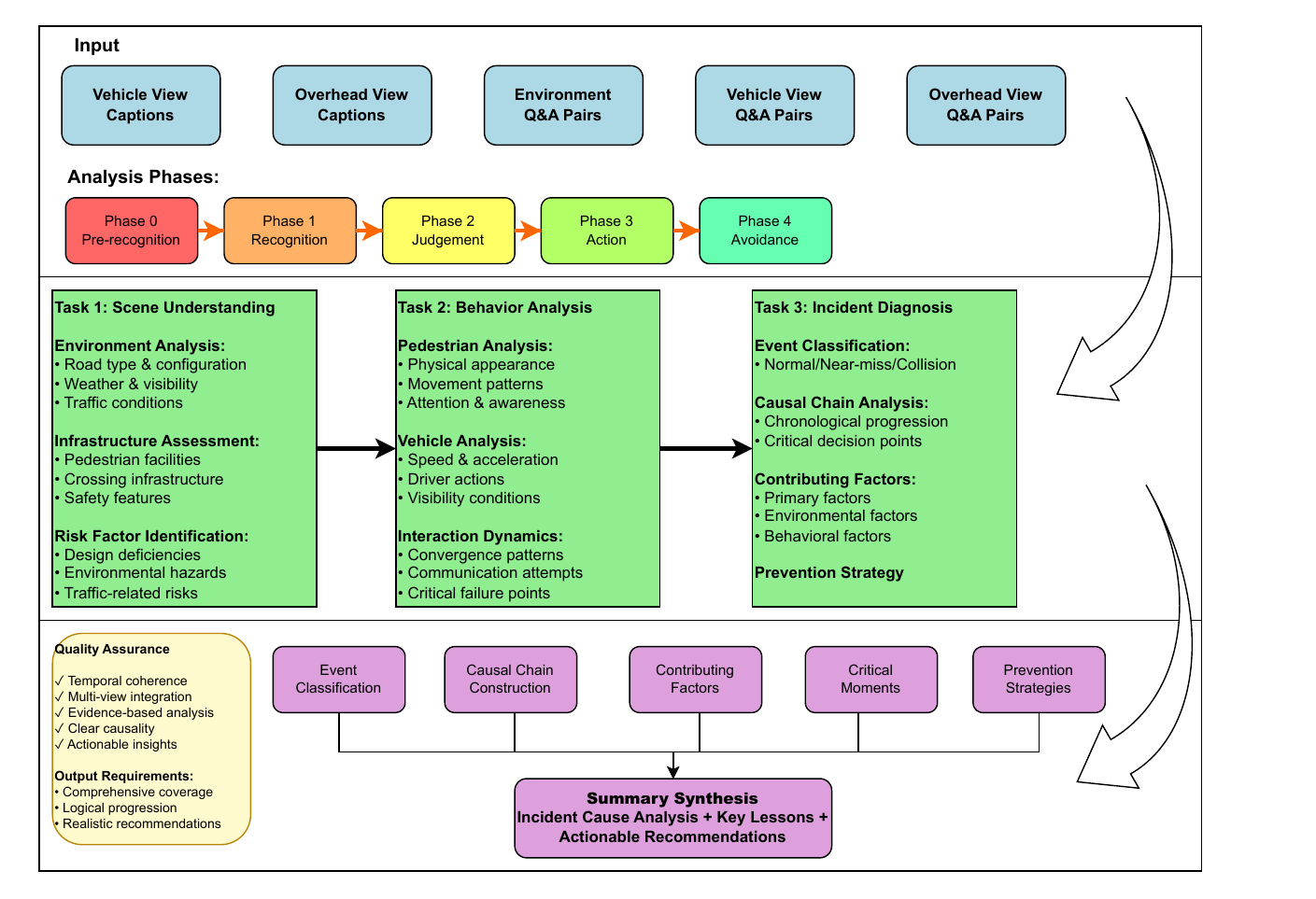}
    \caption{Prompt design logic for Stage 4: Hierarchical Synthesis and Diagnostic Reasoning }
    \label{fig:Prompt design for Stage 4}
\end{figure}

The final stage of the MP-PVIR pipeline focuses on structured synthesis and diagnostic inference. At this stage, the system aggregates the outputs from the previous modules, including segmented phase timelines, phase-specific multi-view captions, and question-answer pairs from both egocentric (vehicle view) and exocentric (overhead view) perspectives. These multi-modal and multi-view textual artifacts serve as the input to a carefully designed prompt that guides a Large Language Model (LLM) through a hierarchical reasoning process.

The prompt is composed of three core components: role conditioning, input formatting, and task instructions, as shown in Figure \ref{fig:Prompt design for Stage 4}. The role specification frames the LLM as a domain expert in pedestrian-vehicle interaction analysis, establishing the expected reasoning depth and analytical lens. The input structure explicitly defines the format and content of the multi-view inputs, encompassing the temporally aligned behavioral phase segments, associated captions, and Q\&A pairs. These inputs are paired with mapping definitions that enable alignment across views and phases. The instruction block decomposes the reasoning task into a progressive sequence that includes scene comprehension, behavior interpretation across actors and viewpoints, causal inference, and diagnostic synthesis.

This stage treats the event as a sequence of behavioral phases grounded in the Perception-Reaction Time paradigm, allowing the LLM to reconstruct the progression of actions and decisions leading to the incident. The model infers incident type, identifies contributing factors, reconstructs the causal chain, and formulates prevention strategies. The result is a comprehensive incident report expressed in a structured JSON format, whose schema is enforced through automatic validation and retry mechanisms. The design ensures coherence, interpretability, and consistency across outputs, thereby enabling scalable, reliable, and cognitively grounded incident analysis.

\section{MP-PVIR Framework Validation and Case studies}
We utilize the pretrained Qwen2.5-VL-7B-Instruct model ~\cite{Qwen2.5-VL} as our backbone for the TG-VLM and PhaVR-VLM. Our training process consists of a single supervised fine-tuning (SFT) stage using parameter-efficient techniques. We employ LoRA (Low-Rank Adaptation) to enable efficient training on limited GPU resources. Video inputs are processed at 2 FPS with a maximum of 6400 pixels to balance computational efficiency and visual quality. We leverage DeepSpeed ZeRO-2 optimization with gradient checkpointing to reduce memory footprint while maintaining training stability. The complete training process takes on 4×A6000 GPUs. The training configuration includes a per-device batch size of 1 with gradient accumulation steps of 64, resulting in an effective batch size of 256 across 4 NVIDIA A6000 GPUs. We train for 6 epochs with a learning rate of 5e-5, utilizing the AdamW optimizer with fused operations and cosine annealing learning rate scheduling with a warm-up ratio of 0.03. Video inputs are processed at 2 FPS with a maximum of 6400 pixels to balance computational efficiency and visual quality. We leverage DeepSpeed ZeRO-2 optimization with gradient checkpointing to reduce memory footprint while maintaining training stability. The model is trained using bfloat16 precision with gradient clipping at a maximum norm of 1.0. 

During inference, we maintain the same video processing parameters (2 FPS, 6400 max pixels) used during training to ensure consistency. The model utilizes the structured prompts for both captioning and question-answering tasks. All evaluations are conducted on the same 4×A6000 GPU configuration, with the complete evaluation process for the full test set.

\subsection{Performance on Pedestrian Behavior Phase Segmentation}
% \begin{table}[h]
% \centering
% \caption{Per-phase mIoU performance comparison for Pedestrian Behavior Phase Segmentation. The overall mIoU for our TG-VLM is 0.4881.}
% \label{tab:temporal_grounding_results}
% \begin{tabular}{l|ccccc}
% \hline
% \textbf{Model} & \textbf{Pre-recog.} & \textbf{Recog.} & \textbf{Judge.} & \textbf{Action} & \textbf{Avoid.} \\
% \hline
% \hline
% Qwen2.5-VL-7b & N/A & N/A & N/A & N/A & N/A \\ %multi-video出不来结果；
% \hline
% Qwen2.5-VL-32b & N/A & N/A & N/A & N/A & N/A \\ %multi-video出不来结果；
% \hline
% \textbf{TG-VLM (Ours)} & \textbf{0.7887} & \textbf{0.5091} & \textbf{0.3662} & \textbf{0.4208} & \textbf{0.3559} \\
% \hline
% \end{tabular}
% \vspace{0.5em} \\
% \small\textit{Note:} N/A indicates that the model is not applicable and failed to produce valid results for multi-stream video inputs.
% \end{table}

Our proposed TG-VLM demonstrates a fundamental capability that the baseline Qwen2.5-VL models entirely lack: the ability to process and reason across multi-view video inputs for pedestrian behavior phase segmentation. While the baseline Qwen2.5-VL (7B and 32B) models failed to produce valid temporal boundaries, our proposed TG-VLM achieved a mean Intersection over Union (mIoU) of 0.4881. The failure of the baseline models stems from two critical gaps: the absence of cognitive phase concepts in their general pre-training data, and the incompatibility of our serialized multi-view input with their single-stream architecture. Without specific fine-tuning, the base models lacked the domain knowledge to ground abstract cognitive states and fails to map visual features across temporally distinct phases. By contrast, our targeted fine-tuning bridged this domain gap, allowing TG-VLM to correctly parse the serialized format and operationalize the behavioral phase taxonomy.

The performance analysis across the sequential behavioral phases reveals a pattern that reflects the evolving complexity of pedestrian-vehicle interactions over time. The model achieves good performance during the Pre-recognition phase, mIoU = 0.7887, when pedestrians have not yet become aware of the approaching vehicle, likely due to the beginning of the whole incident, and less ambiguous behavioral patterns. Performance drops significantly as the interaction progresses through the Recognition phase, with a mIOU of 0.5091, where subtle changes in pedestrian awareness and attention must be detected. The model reaches its low performance during the Judgment phase with an mIOU of 0.3662, representing the brief cognitive moment when pedestrians assess the situation and make critical decisions, a phase characterized by minimal observable visual changes despite profound internal processing. Performance moderately recovers during the Action phase with improved mIOU 0.4208, as physical movements and directional changes provide more observable visual indicators for temporal grounding. Finally, the Avoidance phase, with a mIOU of 0.3559, presents continued challenges, likely due to the rapid and brief duration of these critical moments. This temporal performance progression demonstrates that our model excels at identifying phases with distinct visual markers but faces increasing difficulty as phases become more cognitively driven and temporally compressed.

While the achieved performance establishes feasibility for automated phase segmentation in multi-view settings, the results indicate substantial scope for methodological refinement. While an overall mIoU of 0.4881 indicates room for improvement compared to specialized, task-specific localization models, it represents a significant capability for a Generative VLM. Unlike traditional discriminatory models which output only numerical boundaries, TG-VLM retains the semantic understanding of the scene, allowing for a unified architecture across the pipeline. Future work could explore the integration of pedestrian gaze data, already available in the WTS dataset through synchronized 3D gaze tracking~\cite{kong2024wts}, as an additional input modality for improving phase segmentation.

\subsection{Performance on Phase-Specific Multi-View Videos Reasoning}
To evaluate the reasoning capabilities of PhaVR-VLM independently from segmentation errors, we utilize the ground-truth temporal boundaries for this stage. This isolates the model's ability to interpret visual content from the challenges of temporal localization. 

\subsubsection{Video Captioning Performance}
\begin{table}[htbp]
\centering
\caption{Performance of PhaVR-VLM on Multi-View Video Captioning Tasks}
\label{tab:multiview_captioning}
% \resizebox{\textwidth}{!}{%
\small
\makebox[\textwidth]{%
\begin{tabular}{l|l|ccccc}
\hline
\multirow{2}{*}{\textbf{Model}} & \multirow{2}{*}{\textbf{Dataset/Type}} 
& \multicolumn{5}{c}{\textbf{Video Captioning Metrics}} \\
\cline{3-7}
& & \textbf{BLEU-4} & \textbf{METEOR} & \textbf{ROUGE-L} & \textbf{CIDEr} & \textbf{Score} \\
% \hline
% \hline
% \multicolumn{7}{c}{\cellcolor{gray!20}\textbf{Task 1: Multi-View Video Captioning}} \\
\hline
Qwen-VL-Chat+BE~\cite{duan2024cityllava} & test & 0.243 & 0.451 & 0.439 & 0.692 & 30.03 \\
VideoLLaVA+BE~\cite{duan2024cityllava}   & test & 0.221 & 0.419 & 0.426 & 0.867 & 28.81 \\
Qwen2.5-VL-7b                            & test & N/A   & N/A   & N/A   & N/A   & N/A \\
\hline
\multirow{3}{*}{\textbf{PhaVR-VLM (Ours)}} 
 & test & \textbf{0.292} & \textbf{0.486} & \textbf{0.513} & \textbf{0.315} & \textbf{33.063} \\
 & WTS  & 0.276 & 0.469 & 0.494 & 0.273 & 31.667 \\
 & BDD  & 0.308 & 0.503 & 0.532 & 0.357 & 34.462 \\
\hline
\end{tabular}%
}
\small\textit{Note:} N/A indicates model failure on multi-view video input.
\end{table}

The evaluation results presented in Table~\ref{tab:multiview_captioning} demonstrate the superior performance of our proposed PhaVR-VLM architecture in multi-view video understanding tasks. For the video captioning task, our model achieves significant improvements across all evaluation metrics compared to existing baselines, while addressing a fundamental limitation in current state-of-the-art models—the inability to process multi-view video inputs effectively.

The baseline models from CityLLaVA~\cite{duan2024cityllava}, specifically Qwen-VL-Chat+BE and VideoLLaVA+BE, achieve composite scores of 30.03 and 28.81, respectively, on the test dataset. However, these models are  constrained to single-view video processing, limiting their applicability in complex traffic scenarios requiring multiple perspectives. In stark contrast, our PhaVR-VLM achieves a substantially higher score of 33.063 on the same test dataset while successfully processing synchronized multi-view video streams—a capability that existing models lack entirely.

A critical observation from our experiments is the complete failure of the Qwen2.5-VL-7b model when presented with multi-view video data, as indicated by N/A values across all metrics. This failure underscores the architectural limitations of existing vision-language models and validates the necessity of our specialized fine-tuning approach for multi-view video understanding. The inability of standard VLMs to handle multiple synchronized video streams represents a significant gap in current technology that our approach directly addresses.

Our model demonstrates robust generalization capabilities across different datasets, achieving scores of 31.667 on the WTS dataset and 34.462 on the BDD dataset. The highest performance on the BDD dataset suggests that our approach particularly excels with diverse driving scenarios. When examining individual metrics, PhaVR-VLM shows consistent improvements: BLEU-4 increases from 0.243 from the best baseline to 0.292, METEOR from 0.451 to 0.486, and ROUGE-L from 0.439 to 0.513. These improvements are achieved while handling the substantially more complex task of multi-view video understanding.

\subsubsection{Visual Question Answering Performance}

\begin{table}[htbp]
\centering
\caption{Performance of PhaVR-VLM on Multi-View Visual Question Answering Tasks}
\label{tab:multiview_vqa}
% \resizebox{\textwidth}{!}{%
\small 
\makebox[\textwidth]{%
\begin{tabular}{l|l|cc}
\hline
\textbf{Model} & \textbf{Question Type} & \textbf{Accuracy (\%)} & \textbf{Valid Rate (\%)} \\
% \hline
% \hline
% \multicolumn{4}{c}{\cellcolor{gray!20}\textbf{Task 2: Multi-View Visual Question Answering}} \\
\hline
\multicolumn{4}{l}{\textit{Vehicle View Questions}} \\
\hline
Qwen2.5-VL-7b    & Vehicle View & 45.22 & 93.07 \\
Qwen2.5-VL-32b   & Vehicle View & 49.34 & 97.85 \\
\textbf{PhaVR-VLM (Ours)} & Vehicle View & \textbf{64.70} & \textbf{100} \\
\hline
\multicolumn{4}{l}{\textit{Overhead View Questions}} \\
\hline
Qwen2.5-VL-7b    & Overhead View & 33.52 & 96.55 \\
Qwen2.5-VL-32b   & Overhead View & 37.94 & 97.03 \\
\textbf{PhaVR-VLM (Ours)} & Overhead View & \textbf{50.48} & \textbf{100} \\
\hline
\multicolumn{4}{l}{\textit{Environment Questions}} \\
\hline
Qwen2.5-VL-7b    & Environment & N/A & N/A \\
Qwen2.5-VL-32b   & Environment & N/A & N/A \\
\textbf{PhaVR-VLM (Ours)} & Environment & \textbf{54.44} & \textbf{100} \\
\hline
\end{tabular}%
}
\small\textit{Note:} N/A indicates missing data or unanswerable question category. Valid Rate refers to the percentage of responses in valid multiple-choice format.
\end{table}

As demonstrated in Table \ref{tab:multiview_vqa}, the visual question answering results reveal even more pronounced advantages of our approach, particularly in handling questions that require cross-view reasoning and comprehensive scene understanding. We evaluated model performance across three distinct question categories: Vehicle View, Overhead View, and Environment questions, each presenting unique challenges for multi-view video comprehension.

For Vehicle View questions, our PhaVR-VLM achieves 64.70\% accuracy, representing a substantial improvement of 15.36 percentage points over the stronger Qwen2.5-VL-32b baseline (49.34\%) and 19.48 percentage points over Qwen2.5-VL-7b (45.22\%). This significant performance gap demonstrates the effectiveness of our multi-view fine-tuning approach in capturing vehicle-specific visual features across different camera perspectives.

The Overhead View questions present a different challenge, requiring the model to interpret aerial perspectives that are less common in standard vision-language training data. Here, our model achieves 50.48\% accuracy, outperforming Qwen2.5-VL-32b by 12.54 percentage points and Qwen2.5-VL-7b by 16.96 percentage points. The consistent performance advantage across different viewpoints suggests that our approach successfully learns view-invariant representations while maintaining view-specific understanding capabilities.

Most remarkably, for Environment questions, which require the most comprehensive cross-view reasoning and holistic scene understanding, both baseline models completely fail to produce valid results. In contrast, our PhaVR-VLM successfully handles these complex queries with 54.44\% accuracy. This complete failure of baseline models on Environment questions reveals a critical limitation in existing VLMs: their inability to integrate information across multiple viewpoints for comprehensive scene understanding. The success of our model on these challenging questions validates our architectural design choices and fine-tuning strategy.

An additional noteworthy aspect of our results is the perfect 100\% valid choice rate achieved by PhaVR-VLM across all VQA question categories, compared to the 93.07--97.85\% range observed in baseline models. This improvement in output formatting consistency indicates enhanced instruction-following capabilities, which is crucial for practical deployment in automated traffic analysis systems where reliable, consistently formatted outputs are essential for downstream processing.

In summary, the superior performance of our model across both tasks, video captioning and visual question answering, while maintaining perfect output validity, demonstrates that our multi-view fine-tuning approach not only enables the processing of complex multi-view inputs but also improves the model's overall robustness and reliability. These results establish PhaVR-VLM as a significant advancement in multi-view video understanding for intelligent transportation systems.

\subsection{Hierarchical Synthesis and Diagnostic Reasoning}

This section presents a example analysis of a pedestrian-vehicle collision using multi-view video data and structured annotation. Due to the long output, we summarize the output from Stage 4 into the following aspects.

\subsubsection{Scene Understanding}

Our scene understanding module identified the incident as occurring on a clear, bright day along a straight residential road segment with light traffic. Critically, no sidewalks or designated pedestrian crossings were present, and the flat road design provided no separation between pedestrian and vehicle zones. These infrastructural deficiencies forced pedestrians to walk in the roadway, creating an inherently risky environment.

% \subsubsection{Infrastructure Risk Assessment}

% The scene understanding module identified critical infrastructure deficiencies:

% \begin{itemize}
%     \item \textbf{Primary Risk:} Complete absence of sidewalks forces pedestrians to share the roadway
%     \item \textbf{Separation:} No physical barriers between pedestrian and vehicle zones
%     \item \textbf{Crossing Infrastructure:} No designated pedestrian crossing facilities
%     \item \textbf{Safety Features:} No traffic calming or pedestrian protection measures
% \end{itemize}

% These infrastructure limitations create an environment where pedestrian-vehicle conflicts are inherently more likely, as pedestrians have no choice but to occupy vehicle operational space.

\subsubsection{Appearance and Behavior Analysis}

\begin{table}[htbp]
\centering
\caption{Phase-by-Phase Risk Escalation}
\begin{tabular}{lllll}
\toprule
\textbf{Phase} & \textbf{Time (s)} & \textbf{Pedestrian State} & \textbf{Vehicle Action} & \textbf{Risk Level} \\
\midrule
Pre-recognition & 28.8--29.9 & Standing, distracted & Preparing & Moderate \\
Recognition     & 29.8--30.8 & Still, distracted    & About to reverse & High \\
Judgement       & 30.7--32.5 & Moving forward       & Reversing 5 km/h & High \\
Action          & 32.6--37.8 & In vehicle lane      & Reversing 5 km/h & Critical \\
Avoidance       & 37.8--43.7 & Thrown back          & Collision & Impact \\
\bottomrule
\end{tabular}
\label{tab:phase_risk}
\end{table}

The pedestrian, described as a male in his 30s wearing black clothing and holding a smartphone, remained distracted throughout the interaction. Behavioral tracking across five temporal phases reveals a progressive escalation in risk. Table~\ref{tab:phase_risk} summarizes this trajectory, resulting in a collision.

% \subsubsection{Interaction Dynamics}
The interaction dynamics shown in the Figure \ref{fig:interactionsynamics_json} reveal no observable communication or mutual awareness between the pedestrian and vehicle. The collision was driven by two converging paths: the vehicle reversing and the pedestrian walking forward in its blind zone. The critical failure stemmed from smartphone-induced distraction and a lack of reverse-safety measures on the vehicle.

\begin{figure}[htbp!]
\centering
\begin{mdframed}[linewidth=0.8pt, roundcorner=3pt, backgroundcolor=gray!2]
\begin{verbatim}
"interaction_dynamics": {
  "initial_separation": "Near distance with pedestrian behind vehicle",
  "convergence_pattern": "Vehicle reversing while pedestrian walked forward",
  "communication": "No communication attempts observed",
  "mutual_awareness": "Neither party aware of the other until impact",
  "critical_failure": "Pedestrian smartphone distraction combined with 
                      vehicle reversing"
}
\end{verbatim}
\end{mdframed}
\caption{Model output on interaction dynamics in JSON format}
\label{fig:interactionsynamics_json}
\end{figure}

\subsubsection{Incident Diagnosis and Reasoning}

\begin{figure}[htbp!]
\centering
\begin{mdframed}[linewidth=0.8pt, roundcorner=3pt, backgroundcolor=gray!2]
\begin{verbatim}
{
  "event_diagnosis": {
    "classification": "Collision",
    "severity": "Pedestrian thrown back and fell - potential injury",
    "causal_chain": [
      {
        "phase": "0",
        "factor": "Pedestrian standing behind vehicle while using smartphone"
      },
      {
        "phase": "1", 
        "factor": "Vehicle prepared to reverse while pedestrian remained distracted"
      },
      {
        "phase": "2",
        "factor": "Pedestrian began walking forward while vehicle started reversing"
      },
      {
        "phase": "3",
        "factor": "Both parties converged on collision path with no awareness"
      },
      {
        "phase": "4",
        "factor": "Collision occurred before either party could react"
      }
    ],
    "contributing_factors": {
      "primary": [
        "Pedestrian distraction due to smartphone use",
        "Vehicle reversing without ensuring clear path",
        "Lack of mutual awareness between parties"
      ],
      "environmental": [
        "No sidewalks forcing pedestrian to use vehicle lane",
        "Flat pedestrian-vehicle division with no barriers",
        "Residential road with shared space design"
      ]
    }
  }
}
\end{verbatim}
\end{mdframed}
\caption{Model Output on Incident diagnosis with possible causal chain and contributing factors}
\label{fig:diagnosis_json}
\end{figure}

\FloatBarrier
The diagnostic module synthesizes the scene and behavior analysis to determine causation and prevention strategies, shown in Figure~\ref{fig:diagnosis_json}. It shows that the incident represents a preventable collision caused by the intersection of human factors (smartphone distraction), vehicle operation (reversing in shared space), and infrastructure deficiency (absence of sidewalks). The analysis system successfully identified the multi-factor causation and generated targeted prevention strategies addressing each contributing element.

\section{Conclusion and Discussion}
This study addresses a critical cognitive gap in data-driven traffic safety analytics: the disconnect between pixel-level detection and semantic incident reasoning. Traditional systems can identify when an incident occurs but fail to explain why. To bridge this gap, we propose the Multi-view Phase-aware Pedestrian-Vehicle Incident Reasoning (MP-PVIR) framework, a novel pipeline that transforms raw, multi-stream video data into structured, actionable diagnostic reports.

By integrating classical accident causation theory with state-of-the-art Large Video-Language Models (LVLMs), MP-PVIR moves beyond binary event classification. The framework operationalizes the concept of behavioral phases, deconstructing incidents into pre-recognition, recognition, judgment, action, and avoidance. Our technical approach overcomes the inherent limitations of standard pre-trained VLMs, which are typically restricted to single-view processing. Through targeted fine-tuning, we developed two specialized models: TG-VLM, which successfully grounds temporal phases across synchronized streams, and PhaVR-VLM, which performs deep, cross-view visual reasoning.

Empirical evaluation on the Woven Traffic Safety (WTS) dataset confirms the efficacy of this approach. Where baseline foundation models failed to produce valid outputs for multi-view inputs, our TG-VLM achieved a mean IoU of 0.4881, proving the feasibility of automated behavioral segmentation. Furthermore, PhaVR-VLM demonstrated superior scene understanding capabilities, outperforming baselines in dense captioning (Score: 33.063) and achieving up to 64.70\% accuracy in vehicle-view question answering. Importantly, the hierarchical synthesis stage shows that an LLM agent can aggregate these outputs into human-readable reports, identifying potential causal chains and suggesting targeted prevention strategies.

The implications of this work extend to the domain of Vehicle-Infrastructure Cooperative Systems (VICS) and urban planning. By providing interpretable, phase-by-phase diagnostics, MP-PVIR empowers stakeholders to move from reactive statistics to proactive safety management. Traffic engineers can utilize these insights to identify specific infrastructure deficiencies (e.g., lack of sidewalks causing "Judgment" phase failures) or recurring behavioral risk patterns, supporting data-driven interventions essential for Vision Zero initiatives.

Despite these advancements, several avenues remain for future research. First, the current pipeline is sequential, meaning errors in the initial temporal grounding (TG-VLM) can propagate to the reasoning stage. Future work could explore end-to-end joint training to reduce such cascading errors. Second, while the model performs well on the WTS dataset, generalization to ``long-tail'' edge cases, such as extreme weather, heavy occlusion, or erratic behaviors not represented in training data, remains a challenge. Finally, incorporating additional modalities, specifically pedestrian gaze and pose estimation, could further enhance the granularity of cognitive phase segmentation.

In conclusion, MP-PVIR establishes a new paradigm for AI-empowered traffic safety. By enabling machines to reason about the temporal and causal dynamics of pedestrian-vehicle interactions, this work marks a substantial step toward smart transportation systems that not only observe but understand road user behavior to help prevent fatalities.

\section*{Author Contributions}

The authors confirm contribution to the paper as follows: study conception and design: Hao Zhen, Jidong J. Yang; data collection: Hao Zhen; methodology \& experiments: Hao Zhen; analysis and interpretation of results: Hao Zhen, Jidong J. Yang; manuscript preparation: Hao Zhen, Yunxiang Yang, Jidong J. Yang. All authors reviewed the results and approved the final version of the manuscript.

\newpage

\bibliographystyle{unsrt}  
\bibliography{references}

@techreport{who2023global,
  title={Global status report on road safety 2023},
  author={{World Health Organization}},
  year={2023},
  institution={WHO},
  url={https://www.who.int/publications/i/item/9789240086517}
}

@inproceedings{yu2020bdd100k,
  title={Bdd100k: A diverse driving dataset for heterogeneous multitask learning},
  author={Yu, Fisher and Chen, Haofeng and Wang, Xin and Xian, Wenqi and Chen, Yingying and Liu, Fangchen and Madhavan, Vashisht and Darrell, Trevor},
  booktitle={Proceedings of the IEEE/CVF conference on computer vision and pattern recognition},
  pages={2636--2645},
  year={2020}
}

@inproceedings{kong2024wts,
  title={Wts: A pedestrian-centric traffic video dataset for fine-grained spatial-temporal understanding},
  author={Kong, Quan and Kawana, Yuki and Saini, Rajat and Kumar, Ashutosh and Pan, Jingjing and Gu, Ta and Ozao, Yohei and Opra, Balazs and Sato, Yoichi and Kobori, Norimasa},
  booktitle={European Conference on Computer Vision},
  pages={1--18},
  year={2024},
  organization={Springer}
}

@inproceedings{zhang2023video,
  title={Video-LLaMA: An Instruction-tuned Audio-Visual Language Model for Video Understanding},
  author={Zhang, Hang and Li, Xin and Bing, Lidong},
  booktitle={Proceedings of the 2023 Conference on Empirical Methods in Natural Language Processing: System Demonstrations},
  pages={543--553},
  year={2023}
}

@inproceedings{maazi2024video,
  title={Video-ChatGPT: Towards Detailed Video Understanding via Large Vision and Language Models},
  author={Maazi, Muhammad and Rasheed, Hanoona and Khan, Salman and Khan, Fahad},
  booktitle={62nd Annual Meeting of the Association-for-Computational-Linguistics (ACL)/Student Research Workshop (SRW), Bangkok, THAILAND, aug 11-16, 2024},
  pages={12585--12602},
  year={2024},
  organization={ASSOC COMPUTATIONAL LINGUISTICS-ACL}
}

@inproceedings{dinh2024trafficvlm,
  title={Trafficvlm: A controllable visual language model for traffic video captioning},
  author={Dinh, Quang Minh and Ho, Minh Khoi and Dang, Anh Quan and Tran, Hung Phong},
  booktitle={Proceedings of the IEEE/CVF Conference on Computer Vision and Pattern Recognition},
  pages={7134--7143},
  year={2024}
}

@article{zhang2025language,
  title={When language and vision meet road safety: leveraging multimodal large language models for video-based traffic accident analysis},
  author={Zhang, Ruixuan and Wang, Beichen and Zhang, Juexiao and Bian, Zilin and Feng, Chen and Ozbay, Kaan},
  journal={Accident Analysis \& Prevention},
  volume={219},
  pages={108077},
  year={2025},
  publisher={Elsevier}
}

@inproceedings{duan2024cityllava,
  title={Cityllava: Efficient fine-tuning for vlms in city scenario},
  author={Duan, Zhizhao and Cheng, Hao and Xu, Duo and Wu, Xi and Zhang, Xiangxie and Ye, Xi and Xie, Zhen},
  booktitle={Proceedings of the IEEE/CVF Conference on Computer Vision and Pattern Recognition},
  pages={7180--7189},
  year={2024}
}

@misc{claude_opus_4_2025,
  author = {Anthropic},
  title = {Claude Opus 4},
  year = {2025},
  howpublished = {\url{https://www.anthropic.com}},
  note = {Large Language Model}
}

@inproceedings{guo2025vtg,
  title={Vtg-llm: Integrating timestamp knowledge into video llms for enhanced video temporal grounding},
  author={Guo, Yongxin and Liu, Jingyu and Li, Mingda and Cheng, Dingxin and Tang, Xiaoying and Sui, Dianbo and Liu, Qingbin and Chen, Xi and Zhao, Kevin},
  booktitle={Proceedings of the AAAI Conference on Artificial Intelligence},
  volume={39},
  number={3},
  pages={3302--3310},
  year={2025}
}

@article{Qwen2.5-VL,
  title={Qwen2.5-VL Technical Report},
  author={Bai, Shuai and Chen, Keqin and Liu, Xuejing and Wang, Jialin and Ge, Wenbin and Song, Sibo and Dang, Kai and Wang, Peng and Wang, Shijie and Tang, Jun and Zhong, Humen and Zhu, Yuanzhi and Yang, Mingkun and Li, Zhaohai and Wan, Jianqiang and Wang, Pengfei and Ding, Wei and Fu, Zheren and Xu, Yiheng and Ye, Jiabo and Zhang, Xi and Xie, Tianbao and Cheng, Zesen and Zhang, Hang and Yang, Zhibo and Xu, Haiyang and Lin, Junyang},
  journal={arXiv preprint arXiv:2502.13923},
  year={2025}
}

@article{yang2024enhancing,
  title={Enhancing nighttime vehicle detection with day-to-night style transfer and labeling-free augmentation},
  author={Yang, Yunxiang and Zhen, Hao and Huang, Yongcan and Yang, Jidong J},
  journal={arXiv preprint arXiv:2412.16478},
  year={2024}
}

@inproceedings{chakraborty2018freeway,
  title={Freeway traffic incident detection from cameras: A semi-supervised learning approach},
  author={Chakraborty, Pranamesh and Sharma, Anuj and Hegde, Chinmay},
  booktitle={2018 21st International Conference on Intelligent Transportation Systems (ITSC)},
  pages={1840--1845},
  year={2018},
  organization={IEEE}
}

@article{cambon2009pedestrian,
  title   = {Pedestrian crossing decision-making: A situational and behavioral approach},
  author  = {Cambon de Lavalette, Brigitte and Tijus, Charles and Poitrenaud, S{\'e}bastien and Leproux, Christine and Bergeron, Jacques and Thouez, Jean-Paul},
  journal = {Safety Science},
  volume  = {47},
  number  = {9},
  pages   = {1248--1253},
  year    = {2009},
  publisher = {Elsevier}
}

@article{papadimitriou2009critical,
  title   = {A critical assessment of pedestrian behaviour models},
  author  = {Papadimitriou, Eleonora and Yannis, George and Golias, John},
  journal = {Transportation Research Part F: Traffic Psychology and Behaviour},
  volume  = {12},
  number  = {3},
  pages   = {242--255},
  year    = {2009},
  publisher = {Elsevier}
}

@article{saxena2017analysis,
  title   = {Analysis of Road Traffic Accident using Causation Theory with Traffic Safety Model and Measures},
  author  = {Saxena, Neeta},
  journal = {International Journal for Research in Applied Science \& Engineering Technology},
  volume  = {5},
  number  = {8},
  pages   = {1263--1267},
  year    = {2017}
}

@article{das2021fatal,
  title   = {Fatal pedestrian crashes at intersections: Trend mining using association rules},
  author  = {Das, Subasish and Tamakloe, Reuben and Zubaidi, Hamsa and Obaid, Ihsan and Alnedawi, Ali},
  journal = {Accident Analysis and Prevention},
  volume  = {160},
  pages   = {106306},
  year    = {2021},
  publisher = {Elsevier}
}

@article{miranda2011link,
  title   = {The link between built environment, pedestrian activity and pedestrian--vehicle collision occurrence at signalized intersections},
  author  = {Miranda-Moreno, Luis F. and Morency, Patrick and El-Geneidy, Ahmed M.},
  journal = {Accident Analysis and Prevention},
  volume  = {43},
  number  = {5},
  pages   = {1624--1634},
  year    = {2011},
  publisher = {Elsevier}
}

@article{ni2016evaluation,
  title   = {Evaluation of pedestrian safety at intersections: A theoretical framework based on pedestrian-vehicle interaction patterns},
  author  = {Ni, Ying and Wang, Menglong and Sun, Jian and Li, Keping},
  journal = {Accident Analysis and Prevention},
  volume  = {96},
  pages   = {118--129},
  year    = {2016},
  publisher = {Elsevier}
}

@article{ismail2009automated,
  title   = {Automated analysis of pedestrian--vehicle conflicts using video data},
  author  = {Ismail, Karim and Sayed, Tarek and Saunier, Nicolas and Lim, Clark},
  journal = {Transportation Research Record: Journal of the Transportation Research Board},
  volume  = {2140},
  pages   = {44--54},
  year    = {2009},
  publisher = {Transportation Research Board}
}

@article{sucha2017pedestrian,
  title   = {Pedestrian-driver communication and decision strategies at marked crossings},
  author  = {Sucha, Matus and Dostal, Daniel and Risser, Ralf},
  journal = {Accident Analysis and Prevention},
  volume  = {102},
  pages   = {41--50},
  year    = {2017},
  publisher = {Elsevier}
}

@article{bella2017driver,
  title   = {Driver-pedestrian interaction under different road environments},
  author  = {Bella, Francesco and Natale, Valentina and Silvestri, Manuel},
  journal = {Transportation Research Procedia},
  volume  = {27},
  pages   = {148--155},
  year    = {2017},
  publisher = {Elsevier}
}

@article{tamagusko2022deep,
  title   = {Deep Learning Applied to Road Accident Detection with Transfer Learning and Synthetic Images},
  author  = {Tamagusko, Tiago and Correia, Matheus Gomes and Huynh, Minh Anh and Ferreira, Adelino},
  journal = {Transportation Research Procedia},
  volume  = {64},
  pages   = {90--97},
  year    = {2022},
  publisher = {Elsevier}
}

@article{zhang2023traffic,
  title   = {Traffic Accident Detection Method Using Trajectory Tracking and Influence Maps},
  author  = {Zhang, Yihang and Sung, Yunsick},
  journal = {Mathematics},
  volume  = {11},
  number  = {7},
  pages   = {1743},
  year    = {2023},
  publisher = {MDPI}
}

@article{zahid2024datadriven,
  title   = {A Data-Driven Approach for Road Accident Detection in Surveillance Videos},
  author  = {Zahid, Ariba and Qasim, Tehreem and Bhatti, Naeem and Zia, Muhammad},
  journal = {Multimedia Tools and Applications},
  volume  = {83},
  pages   = {17217--17231},
  year    = {2024},
  publisher = {Springer}
}

@article{li2021attention,
  title   = {Crash Report Data Analysis for Creating Scenario-Wise, Spatio-Temporal Attention Guidance to Support Computer Vision-Based Perception of Fatal Crash Risks},
  author  = {Li, Yu and Karim, Muhammad Monjurul and Qin, Ruwen and Sun, Zeyi and Wang, Zuhui and Yin, Zhaozheng},
  journal = {Accident Analysis and Prevention},
  volume  = {151},
  pages   = {105962},
  year    = {2021},
  publisher = {Elsevier}
}

@article{adewopo2024smart,
  title   = {Smart City Transportation: Deep Learning Ensemble Approach for Traffic Accident Detection},
  author  = {Adewopo, Victor A. and Elsayed, Nelly},
  journal = {IEEE Access},
  volume  = {12},
  pages   = {59134--59149},
  year    = {2024},
  publisher = {IEEE}
}

@article{liu2022smart,
  title   = {Smart Traffic Monitoring System Using Computer Vision and Edge Computing},
  author  = {Liu, Guanxiong and Shi, Hang and Kiani, Abbas and Khreishah, Abdallah and Lee, Joyoung and Ansari, Nirwan and Liu, Chengjun and Yousef, Mustafa Mohammad},
  journal = {IEEE Transactions on Intelligent Transportation Systems},
  volume  = {23},
  number  = {8},
  pages   = {12027--12040},
  year    = {2022},
  publisher = {IEEE}
}

@article{aboah202xvision,
  title   = {A Vision-Based System for Traffic Anomaly Detection Using Deep Learning and Decision Trees},
  author  = {Aboah, Armstrong and Shoman, Maged and Mandal, Vishal and Davami, Sayedomidreza and Adu-Gyamfi, Yaw and Sharma, Anuj},
  journal = {Technical report},
  year    = {202x},
  note    = {Unpublished manuscript; please update venue and year when available}
}

@article{fang2024visiontad,
  title   = {Vision-Based Traffic Accident Detection and Anticipation: A Survey},
  author  = {Fang, Jianwu and Qiao, Jiahuan and Xue, Jianru and Li, Zhengguo},
  journal = {IEEE Transactions on Circuits and Systems for Video Technology},
  volume  = {34},
  number  = {4},
  pages   = {1983--2009},
  year    = {2024},
  publisher = {IEEE}
}

@article{basheer2023realtime,
  title   = {A Real-Time Computer Vision Based Approach to Detection and Classification of Traffic Incidents},
  author  = {Basheer Ahmed, Mohammed Imran and Zaghdoud, Rim and Ahmed, Mohammed Salih and Sendi, Razan and Alsharif, Sarah and Alabdulkarim, Jomana and Albin Saad, Bashayr Adnan and Alsabt, Reema and Rahman, Atta and Krishnasamy, Gomathi},
  journal = {Big Data and Cognitive Computing},
  volume  = {7},
  number  = {1},
  pages   = {22},
  year    = {2023},
  publisher = {MDPI}
}

@inproceedings{ijjina2019computervision,
  title     = {Computer Vision-Based Accident Detection in Traffic Surveillance},
  author    = {Ijjina, Earnest Paul and Chand, Dhananjai and Gupta, Savyasachi and Goutham, K.},
  booktitle = {Proceedings of the 10th International Conference on Computing, Communication and Networking Technologies (ICCCNT)},
  year      = {2019},
  organization = {IEEE},
  address   = {Kanpur, India}
}

@inproceedings{munasinghe2025videoglamm,
  title        = {VideoGLaMM: A Large Multimodal Model for Pixel-Level Visual Grounding in Videos},
  author       = {Munasinghe, Shehan and Gani, Hanan and Zhu, Wenqi and Cao, Jiale and Xing, Eric and Khan, Fahad Shahbaz and Khan, Salman},
  booktitle    = {Proceedings of the IEEE/CVF Conference on Computer Vision and Pattern Recognition (CVPR)},
  year         = {2025},
  organization = {IEEE}
}

@article{wang2025groundedvideollm,
  title   = {Grounded-VideoLLM: Sharpening Fine-grained Temporal Grounding in Video Large Language Models},
  author  = {Wang, Haibo and Xu, Zhiyang and Cheng, Yu and Diao, Shizhe and Zhou, Yufan and Cao, Yixin and Ge, Weifeng and Huang, Lifu},
  journal = {arXiv preprint arXiv:2410.03290},
  year    = {2025}
}

@inproceedings{qu2024chatvtg,
  title        = {ChatVTG: Video Temporal Grounding via Chat with Video Dialogue Large Language Models},
  author       = {Qu, Mengxue and Chen, Xiaodong and Liu, Wu and Li, Alicia and Zhao, Yao},
  booktitle    = {Proceedings of the IEEE/CVF Conference on Computer Vision and Pattern Recognition Workshops (CVPRW)},
  year         = {2024},
  organization = {IEEE}
}

@inproceedings{bossen2025gestvlm,
  title        = {Can Vision--Language Models Understand and Interpret Dynamic Gestures from Pedestrians? Pilot Datasets and Exploration Towards Instructive Nonverbal Commands for Cooperative Autonomous Vehicles},
  author       = {Bossen, Tonko and M{\o}gelmose, Andreas and Greer, Ross},
  booktitle    = {Proceedings of the IEEE/CVF Conference on Computer Vision and Pattern Recognition Workshops (CVPRW)},
  year         = {2025},
  organization = {IEEE}
}

@article{zhang2025seeunsafe,
  title   = {When Language and Vision Meet Road Safety: Leveraging Multimodal Large Language Models for Video-based Traffic Accident Analysis},
  author  = {Zhang, Ruixuan and Wang, Beichen and Zhang, Juexiao and Bian, Zilin and Feng, Chen and Ozbay, Kaan},
  journal = {Accident Analysis and Prevention},
  volume  = {219},
  pages   = {108077},
  year    = {2025},
  publisher = {Elsevier}
}

@article{teng2025vlm_crossing,
  title   = {Improving Intelligent Perception and Decision Optimization of Pedestrian Crossing Scenarios in Autonomous Driving Environments Through Large Visual Language Models},
  author  = {Teng, Xiao and Huang, Lin and Shen, Zhenjiang and Li, Wankai and others},
  journal = {Scientific Reports},
  volume  = {15},
  number  = {31283},
  year    = {2025},
  publisher = {Nature Publishing Group}
}

@article{heinrich1941industrial,
  title={Industrial Accident Prevention. A Scientific Approach.},
  author={Heinrich, Herbert William},
  year={1941}
}

@book{reason1990human,
  title={Human error},
  author={Reason, James},
  year={1990},
  publisher={Cambridge university press}
}

@article{khanam2024yolov5,
  title={What is YOLOv5: A deep look into the internal features of the popular object detector},
  author={Khanam, Rahima and Hussain, Muhammad},
  journal={arXiv preprint arXiv:2407.20892},
  year={2024}
}

@inproceedings{wojke2017simple,
  title={Simple online and realtime tracking with a deep association metric},
  author={Wojke, Nicolai and Bewley, Alex and Paulus, Dietrich},
  booktitle={2017 IEEE international conference on image processing (ICIP)},
  pages={3645--3649},
  year={2017},
  organization={IEEE}
}

@inproceedings{reulke2007traffic,
  title={Traffic surveillance using multi-camera detection and multi-target tracking},
  author={Reulke, Ralf and Bauer, Sascha and Doring, T and Meysel, Frederik},
  booktitle={Image and Vision Computing New Zealand},
  pages={175--180},
  year={2007}
}

@inproceedings{hsu2020traffic,
  title={Traffic-aware multi-camera tracking of vehicles based on reid and camera link model},
  author={Hsu, Hung-Min and Wang, Yizhou and Hwang, Jenq-Neng},
  booktitle={Proceedings of the 28th ACM International Conference on Multimedia},
  pages={964--972},
  year={2020}
}

@article{zhen2025tab,
  title={Tab-text: Bridging Tabular Data and Natural Language for Enhanced Traffic Safety Analysis and Modeling},
  author={Zhen, Hao and Yang, Jidong J},
  journal={Expert Systems with Applications},
  pages={128450},
  year={2025},
  publisher={Elsevier}
}

@article{zhen2024leveraging,
  title={Leveraging large language models with chain-of-thought and prompt engineering for traffic crash severity analysis and inference},
  author={Zhen, Hao and Shi, Yucheng and Huang, Yongcan and Yang, Jidong J and Liu, Ninghao},
  journal={Computers},
  volume={13},
  number={9},
  pages={232},
  year={2024},
  publisher={MDPI}
}

@article{ZHEN2025100030,
title = {CrashSage: A large language model-centered framework for contextual and interpretable traffic crash analysis},
journal = {Artificial Intelligence for Transportation},
volume = {3-4},
pages = {100030},
year = {2025},
issn = {3050-8606},
doi = {https://doi.org/10.1016/j.ait.2025.100030},
url = {https://www.sciencedirect.com/science/article/pii/S3050860625000304},
author = {Hao Zhen and Jidong J. Yang}
}

@phdthesis{zhen2025bridging,
  title={Bridging Structure and Semantics: A Multimodal Interpretable LLM Framework for Traffic Crash Modeling and Analysis},
  author={Zhen, Hao},
  year={2025},
  school={University of Georgia}
}

@article{zhao2025auto,
  title={How to Auto-optimize Prompts for Domain Tasks? Adaptive Prompting and Reasoning through Evolutionary Domain Knowledge Adaptation},
  author={Zhao, Yang and Wang, Pu and Yang, Hao Frank},
  journal={arXiv preprint arXiv:2510.21148},
  year={2025}
}

@inproceedings{gorrini2017crossing,
  title={Crossing behaviour of social groups: Insights from observations at non-signalised intersection},
  author={Gorrini, Andrea and Crociani, Luca and Vizzari, Giuseppe and Bandini, Stefania},
  booktitle={International Conference on Traffic and Granular Flow},
  pages={443--450},
  year={2017},
  organization={Springer}
}

@incollection{gorrini2016towards,
  title={Towards Modeling Pedestrian-Vehicle Interactions: Empirical Study on Urban Unsignalized Intersection},
  author={Gorrini, A and Vizzari, G and Bandini, S and others},
  booktitle={Proceedings of Pedestrian and Evacuation Dynamics 2016},
  year={2016}
}

@article{yang2025structured,
  title={Structured Prompting and Collaborative Multi-Agent Knowledge Distillation for Traffic Video Interpretation and Risk Inference},
  author={Yang, Yunxiang and Xu, Ningning and Yang, Jidong J},
  journal={Computers},
  volume={14},
  number={11},
  pages={490},
  year={2025},
  publisher={MDPI}
}

@article{yang2025multi,
  title={Multi-Agent Visual-Language Reasoning for Comprehensive Highway Scene Understanding},
  author={Yang, Yunxiang and Xu, Ningning and Yang, Jidong J},
  journal={arXiv preprint arXiv:2508.17205},
  year={2025}
}

@article{zhen2022co,
  title={Co-supervised learning paradigm with conditional generative adversarial networks for sample-efficient classification},
  author={Zhen, Hao and Shi, Yucheng and Yang, Jidong J and Vehni, Javad Mohammadpour},
  journal={arXiv preprint arXiv:2212.13589},
  year={2022}
}

@article{yang2023strategic,
  title={Strategic Sensor Placement in Expansive Highway Networks: A Novel Framework for Maximizing Information Gain},
  author={Yang, Yunxiang and Yang, Jidong J},
  journal={Systems},
  volume={11},
  number={12},
  pages={577},
  year={2023},
  publisher={MDPI}
}

@article{yang2024information,
  title={An Information Gradient Approach to Optimizing Traffic Sensor Placement in Statewide Networks.},
  author={Yang, Yunxiang and Zhen, Hao and Yang, Jidong J},
  journal={Information (2078-2489)},
  volume={15},
  number={10},
  year={2024}
}

\end{document}